\documentclass[conference]{IEEEtran}
\IEEEoverridecommandlockouts
\usepackage{cite}
\usepackage{amsmath,amssymb,amsfonts}
\usepackage{algorithmic}
\usepackage{graphicx}
\usepackage{textcomp}
\usepackage{xcolor}
\usepackage{amssymb}
\usepackage{booktabs}
\usepackage{multirow} 
\usepackage{setspace}

\usepackage{url}
\usepackage{hyperref} 
\usepackage{array}
\usepackage{stfloats}
\usepackage{verbatim}
\usepackage{subcaption}
\usepackage{caption}
\usepackage{tabularx}
\usepackage{lipsum}   
\usepackage{multirow} 

\def\BibTeX{{\rm B\kern-.05em{\sc i\kern-.025em b}\kern-.08em
    T\kern-.1667em\lower.7ex\hbox{E}\kern-.125emX}}

\begin{document}

\title{RefSAM: Efficiently Adapting Segmenting Anything Model for Referring Video Object Segmentation
}

\author{\IEEEauthorblockN{
Yonglin Li\textsuperscript{1}, 
Jing Zhang\textsuperscript{2}, 
Xiao Teng\textsuperscript{1},
Long Lan\textsuperscript{1}\IEEEauthorrefmark{1}\thanks{\IEEEauthorrefmark{1}Corresponding author.},
Xinwang Liu\textsuperscript{1},
}
\IEEEauthorblockA{
\textsuperscript{1}College of Computer Science and Technology, National University of Defense Technology, Changsha, China 410073\\
\textsuperscript{2}School of Computer Science, The University of Sydney, Sydney, Australia 2008\\
}}

\maketitle

\begin{abstract}
The Segment Anything Model (SAM) has gained significant attention for its impressive performance in image segmentation. However, it lacks proficiency in referring video object segmentation (RVOS) due to the need for precise user-interactive prompts and a limited understanding of different modalities, such as language and vision. This paper presents the RefSAM model, which explores the potential of SAM for RVOS by incorporating multi-view information from diverse modalities and successive frames at different timestamps in an online manner. Our proposed approach adapts the original SAM model to enhance cross-modality learning by employing a lightweight Cross-Modal MLP that projects the text embedding of the referring expression into sparse and dense embeddings, serving as user-interactive prompts. Additionally, we have introduced the hierarchical dense attention module to fuse hierarchical visual semantic information with sparse embeddings to obtain fine-grained dense embeddings, and an implicit tracking module to generate a tracking token and provide historical information for the mask decoder.
Furthermore, we employ a parameter-efficient tuning strategy to align and fuse the language and vision features effectively. Through comprehensive ablation studies, we demonstrate our model's practical and effective design choices. Extensive experiments conducted on Refer-Youtube-VOS, Ref-DAVIS17, and three referring image segmentation datasets validate the superiority and effectiveness of our RefSAM model over existing methods. The code and models will be made publicly at \href{https://github.com/LancasterLi/RefSAM}{github.com/LancasterLi/RefSAM}.
\end{abstract}

\begin{IEEEkeywords}
Video Object Segmentation, Vision Transformer, Language and Vision, Segment Anything
\end{IEEEkeywords}

\section{Introduction}\label{sec1}
Referring video object segmentation (RVOS) aims to accurately segment the target object in the video with the guidance of given language expressions \cite{botach2022end, miao2023self}. It has attracted widespread attention in the computer vision community due to its great potential in real-world applications, such as video retrieval and video editing \cite{zhang2020empowering,wu2022language,li2023referring}. It is a challenging task and the key lies in how to incorporate multi-view information from video frames at different timestamps and align sources from different modalities, \textit{i.e.}, vision and language views \cite{lu2024zero, he2023vgsg}. Compared with video object segmentation, RVOS is more challenging as it is required to perform semantic-level alignment between vision and language modalities \cite{zhao2024learning, zhang2024multi} without the true mask labeling in the first frame. Although many efforts have been devoted to tackling this task  
\cite{ding2022language, zhao2022modeling, ding2021progressive}, 
it still remains a challenge how to accurately identify and segment the target object in the video by fully exploiting the semantic meanings of sources from vision and language views. 

Recently, the Segment Anything Model (SAM) \cite{kirillov2023segment} has been proposed, which serves as a foundation model for image segmentation. Based on the designed prompt engineering, it can be transferred to the new task in a zero-shot manner. Due to its impressive performance on image segmentation, many works have explored to apply SAM in other related fields~\cite{jing2023segment}, \textit{e.g.}, medical image analysis \cite{yue2024surgicalsam},
remote sensing images \cite{wang2024samrs}, video object tracking \cite{cheng2023segment, yang2023track}, style transfer \cite{liu2023any}, and 3D reconstruction \cite{shen2023anything}. Although these methods have achieved great progress on these related tasks based on SAM, none of them can be applied to RVOS directly due to the inherent complexity of this task. Concretely, RVOS requires the model to have a comprehensive understanding of sources from vision and language modalities. Unlike video object segmentation (VOS), no ground-truth mask annotations in the first frame are available in RVOS. It's worth noting that although some SAM-based multiple object tracking models such as SAM-Track \cite{cheng2023segment} and TAM \cite{yang2023track} can also be adapted to the task of RVOS by providing bounding boxes of the first frame detected by extra object detection models, such as Grounding DINO \cite{liu2023grounding}, some non-negligible issues can exist in such frameworks. For example, they cannot work effectively in an end-to-end manner due to the multi-stage pipeline. As a result, inaccurate bounding boxes in the first frame detected by the object detection model can result in the poor performance of the downstream segmentation task. On the other hand, the involvement of separate models can bring extra difficulties in the model training and deployment process.  

\begin{figure*}[!t]
\centering
\includegraphics[width=0.9\linewidth]{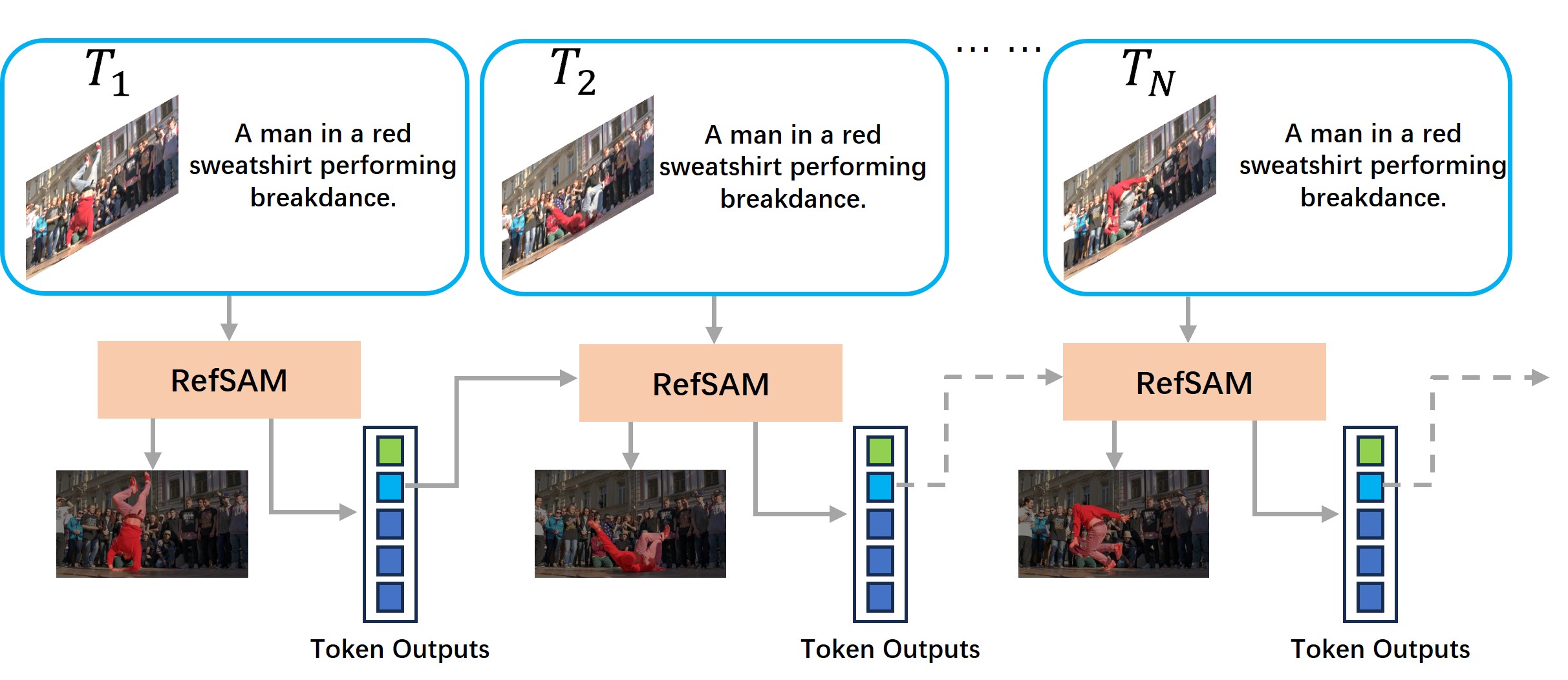}
\caption{RefSAM integrates multi-view information from diverse modalities and successive frames at different timestamps in an online manner. Similar to SAM, RefSAM generates three types of token outputs: IOU token output (green), main mask token output (light blue), and three-scale mask token outputs (dark blue). We feed the main mask token output into RefSAM's tracking module to obtain the track token that can provide historical information for the mask decoder and assist RefSAM in predicting the mask for the next frame.}
\label{fig:track_forward}
\end{figure*}

Thus, a natural question arises that how to effectively adapt SAM to the RVOS task in an end-to-end manner to fully unleash its potential capacity for video segmentation and multimodal fusion. In this paper, we conduct the initial exploration and propose the RefSAM, which is the first end-to-end SAM-based framework for the task of RVOS, as shown in Figure~\ref{fig:track_forward}. Based on the powerful foundation model SAM and the specifically designed modules, RefSAM can perform accurate target object segmentation in the video with given language expressions in an online manner. The overall pipeline of RefSAM as depicted in Figure \ref{fig:overall_network}. Specifically, to enhance the cross-modality learning capability of the original SAM, we propose a lightweight Cross-Modal MLP that projects the text embedding of the referring expression into sparse and dense embeddings, serving as user-interactive prompts like points and bounding boxes prompts. Additionally, we have introduced the hierarchical dense attention module to fuse hierarchical visual semantic information with sparse embeddings in order to obtain fine-grained dense embeddings, and an implicit tracking module to generate a tracking token and provide historical information for the mask decoder. Furthermore, we employ a parameter-efficient tuning strategy to effectively align and fuse the language and vision features.

Our contributions can be concluded as follows. 
\begin{enumerate}
\item[$\bullet$] We conduct the pioneering study to explore SAM's potential for RVOS through the integration of multi-view information from diverse modalities and successive frames at different timestamps. Specifically, we introduce RefSAM, a novel approach that utilizes lightweight modules and an efficient fine-tuning strategy to align and fuse language and vision features in an end-to-end learning manner.

\item[$\bullet$] We design the hierarchical dense attention module for leveraging diverse levels of visual features and textual features to effectively perform cross-modal segmentation of objects with different sizes. We also devise an implicit tracking module to generate a track token and provide historical information to the mask decoder, which effectively enhances the spatiotemporal consistency of segmentation results.

\item[$\bullet$] Extensive experiments on Refer-Youtube-VOS, Ref-DAVIS17, and three Referring image segmentation (RIS) datasets demonstrate the promising results of RefSAM and its superiority compared to existing methods. 

\end{enumerate}


\section{Related Work}\label{sec2}
\subsection{Referring Video Object Segmentation}

\cite{gavrilyuk2018actor} first proposes the RVOS task, and presents an end-to-end pipeline that uses dynamic convolution for mask generation. Additionally, \cite{gavrilyuk2018actor} extends the A2D-sentence \cite{xu2015can} and J-HMDB \cite{jhuang2013towards} datasets with text descriptions. \cite{khoreva2019video} first uses the grounding referring expression method to generate object boxes, and then utilizes a video segmentation model to create masks. \cite{khoreva2019video}
extends the DAVIS16 and DAVIS17 \cite{pont20172017} datasets using two types of annotated language descriptions: ``first frame'' and ``full video''. Later, URVOS \cite{seo2020urvos} presents a large-scale RVOS dataset Refer-Youtube-VOS and proposes a new method that promotes cross-modal interactions. Many methods contribute to leveraging different levels of visual features to improve performance. RefVOS \cite{bellver2020refvos} utilizes the feature from the last stage of the encoder for cross-modal fusion. ClawCraneNet \cite{liang2021clawcranenet} explores the value of object-level information. MRSA \cite{wu2022multi} simultaneously acquires the video-level, frame-level, and object-level information during the encoding stage. Some other methods incorporate grounding techniques to assist in generating segmentation masks, such as CITD \cite{liang2021rethinking} and \cite{ding2021progressive}. The former uses a grounding model for the final segmentation generation, while the latter employs one grounding model to select the best proposal for post-processing. Other approaches explore the utility of motion information. For example, MMVT \cite{zhao2022modeling} uses flow maps to capture motion information, while LBDT \cite{ding2022language} relies on frame difference. Although these models have achieved promising performance, more research efforts are still needed to further address the significant semantic gap between visual information and language information.

Vision Transformers (ViTs)~\cite{dosovitskiy2020image} have attracted lots of attention recently and have been widely used in various 
domains~\cite{liu2021swin,zhang2023vitaev2,wang2022towards,xuvitpose,lan2022siamese,xie2024csfwinformer,li2024binsformer},
including the field of RVOS. MTTR \cite{botach2022end} takes both a video and a text query as input, and it generates prediction sequences for all objects based on this information. It uses cross-entropy to align the video and text during the prediction process. ReferFormer \cite{wu2022language} uses a set of language expression-based object queries as inputs to the Transformer decoder. These queries specifically focus on the referred object, and their purpose is to generate dynamic kernels. VLT \cite{ding2021vision} utilizes the QGM to produce various query vectors. These query vectors play a crucial role in guiding the Transformer decoder throughout the decoding process, ultimately generating appropriate responses. Different from these models, we present a simple yet effective solution by adapting SAM to RVOS via some minimal designs.

\subsection{Segment Anything Model}

Recently, SAM \cite{kirillov2023segment} builds a foundation model for segmentation. The model designs a promptable segmentation task and it can transfer zero-shot to new image distributions and tasks. Following that, a large number of research studies based on SAM emerged. SAM has been widely applied in various fields~\cite{jing2023segment}, \textit{e.g.}, medical image analysis \cite{yue2024surgicalsam}, remote sensing images \cite{wang2024samrs}, video object tracking \cite{cheng2023segment, yang2023track}, style transfer \cite{liu2023any}, and 3D reconstruction \cite{shen2023anything}. In the medical image analysis domain, SAMed \cite{zhang2023customized} adds LoRA layers while freezing the image encoder to tune SAM for adapting to medical images. \cite{putz2023segment} demonstrates that SAM can achieve remarkable zero-shot segmentation accuracy on brain tumor MRI datasets. 
In the remote sensing images domain, \cite{julka2023knowledge} adds a domain decoder in SAM to learn specific features for the remote sensing images, while \cite{wang2024samrs} leverages SAM to generate a large-scale dataset: SAMRS. 
In the style transfer domain, Any-to-Any Style Transfer \cite{liu2023any} uses SAM to extract the segmentation mask and uses the VGG encoder and decoder to transform the segmentation mask into another image. In the 3D reconstruction domain, SA3D \cite{cen2023segment} employs a pre-trained NeRF and SAM to obtain 2D images, and utilizes inverse rendering and cross-view self-prompting to acquire corresponding 3D masks. These advancements have shown the versatility of the foundation segmentation model.

PerSAM \cite{zhang2023personalize} introduces target-guided attention and target-semantic prompting to enhance SAM's performance in one-shot scenarios. Additionally, the model adaptively concatenates masks from three different scales, further improving its segmentation accuracy. HQ-SAM \cite{ke2024segment} incorporates a learnable HQ-token in SAM's mask decoder to enhance the refinement of SAM's output mask. 
SAM-Track \cite{cheng2023segment} and TAM \cite{yang2023track} enable users to select multiple objects in videos for tracking \cite{lan2020semi, tan2021nocal, lan2018interacting}. However, none of these models are currently available for the RVOS task. In this paper, we propose an end-to-end RVOS model based on SAM for the first time.


\section{RefSAM for Reffering Video Object Segmentation}\label{sec3}
In this section, we introduce the RefSAM model. First, we present the preliminaries of SAM in Sec. \ref{sec:Preliminaries: SAM}. Then, in Sec. \ref{sec:Method overview}, we provide an overview of the proposed RefSAM model. 
Detailed descriptions of each component of RefSAM are provided in the subsequent sub-sections.

\begin{figure*}[!t]
\centering
\includegraphics[width=1.01\linewidth]{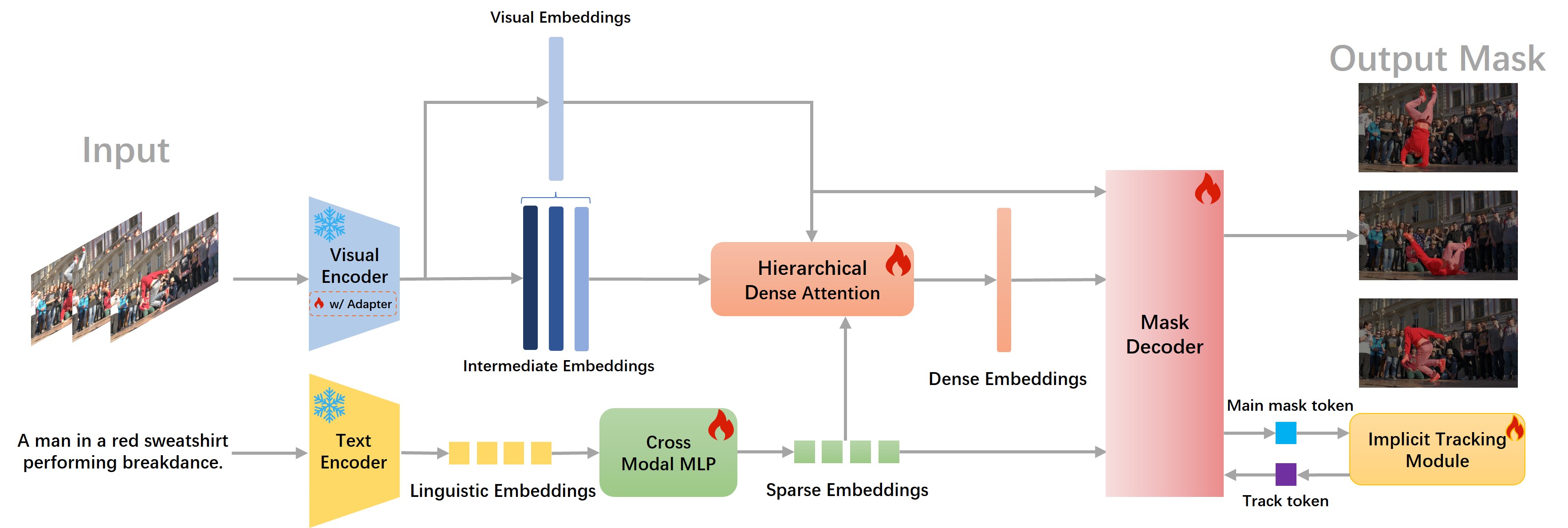}
\caption{The overall pipeline of RefSAM. It mainly consists of five key components: 1) Backbone: Visual Encoder of SAM \cite{kirillov2023segment} with Adapter and Text Encoder; 2) Cross-Modal MLP; 3) Hierarchical Dense Attention; 4) Mask Decoder of SAM; and 5) Implicit Tracking Module. We construct cross-modal Sparse Embeddings and Dense Embeddings to learn text-visual information and predict masks. 
We use the implicit tracking module to generate a track token and provide historical information for the mask decoder.
}
\label{fig:overall_network}
\end{figure*}

\subsection{Preliminaries}
\label{sec:Preliminaries: SAM}
\textbf{Task Definition}: 
The RVOS task leverages both visual and text information. %
Given a video clip $ \mathcal{L}$ $ = \{I_{t}\}_{t=1}^{T}$ with $T$ frames and a referring expression $E = \{ e_{l} \}_{l=1}^{L}$ with $L$ words, the goal of RVOS is to produce $T$-frame binary segmentation masks $S = \{s_{t}\}_{t=1}^{T}$, $s_{t} \in \mathbb{R}^{H \times W}$ of the referred object.

\textbf{Architecture of SAM}: SAM \cite{kirillov2023segment} mainly consists of three components: an image encoder, a prompt encoder, and a mask decoder. The image encoder is a ViT-based \cite{dosovitskiy2020image} backbone to extract image features. The prompt encoder is responsible for encoding two sets of prompts: sparse prompts (consisting of points and boxes) and dense prompts (comprising masks). These prompts contain interactive positional information, which is then provided to the mask decoder for further processing. The mask decoder consists of a two-layer transformer. It takes image embeddings, output tokens, and prompt tokens as inputs and generates three-scale masks along with corresponding IoU scores as outputs. SAM demonstrates strong zero-shot generalization in the segmentation task. However, SAM has limitations in effectively leveraging text for segmentation, and its training process is computationally expensive due to the large scale of the SA-1B dataset and the huge amount of parameters. 

\subsection{Overview of RefSAM's Architecture}
\label{sec:Method overview}

We introduce the RefSAM model to efficiently adapt SAM to the RVOS task and boost the potential segmentation capacity of SAM. As shown in Figure~\ref{fig:overall_network}, it mainly consists of five key components: two types of modality encoders, a Cross-Modal MLP, a Hierarchical Dense Attention (HDA) module, the Mask Decoder, and an Implicit Tracking Module (ITM). 
Firstly, we use the Visual Encoder of SAM to extract frame features as visual embeddings. 
Meanwhile, we use the text-to-text model (T5) \cite{raffel2020exploring} as the Text Encoder to extract linguistic embeddings. Then, we construct cross-modal Sparse Embeddings to provide aligned linguistic information to the Mask Decoder through the Cross-Modal MLP. Next, RefSAM utilizes the HDA module to fuse hierarchical visual semantic information and sparse embeddings in order to obtain fine-grained dense embeddings. Finally, the Mask Decoder of SAM leverages sparse embeddings, dense embeddings, and visual embeddings for final mask prediction and generates mask token output. The ITM utilizes the mask token output to generate a track token, which encodes historical information for the mask decoder. During training, RefSAM reuses the pre-trained weights of SAM and text encoder T5 while only fine-tuning lightweight modules, \textit{i.e.}, Adapter, Cross-Modal MLP, HDA, ITM, and Mask Decoder, enabling parameter-efficient fine-tuning. During inference, we directly output the mask predictions by selecting the masks with the highest score as the final results.

\subsection{Backbone}

\subsubsection{Visual Encoder}

We start by adopting the image encoder of SAM as the visual encoder $\mathcal{E}_{v}$ to extract the visual feature maps and intermediate feature maps for each frame in a video clip, as shown in the blue part in the top of Figure~\ref{fig:overall_network}. The image encoder $\mathcal{E}_{v}$ is an MAE \cite{he2022masked} pre-trained ViT backbone \cite{dosovitskiy2020image}. Specifically, for each frame $I_{t}$ in the video clip $ \mathcal{L}$ $ = \{I_{t}\}_{t=1}^{T}$, the vision encoder $\mathcal{E}_{v}$ is adopted to extract the feature maps Set for this frame, \textit{i.e.}, 
\begin{equation}
    f_{t}, f_t^{mid} = \mathcal{E}_{v}(I_{t}),
\end{equation}
where $f_{t}\in \mathbb{R}^{C_{v} \times H_{0} \times W_{0}}$ is the corresponding visual feature map of frame $I_{t}$ from the last layer of the SAM encoder. And $f_t^{mid} = \{f_{t}^{i}\}_{i=1}^{3}\in \mathbb{R}^{C_{mid} \times H_{0} \times W_{0}}$ denotes the set of intermediate feature maps of frame $I_{t}$. By applying the vision encoder $\mathcal{E}_{v}$ on each frame independently, a set of visual feature maps $ \mathcal{F}_{v} = \{f_{t}\}_{t=1}^{T}$ and a set of intermediate feature maps $ \mathcal{F}_{mid} = \{f^{mid}_{t}\}_{t=1}^{T}$ can be obtained for $T$ frames in the video clip. As SAM has strong zero-shot segmentation performance, we freeze the parameters of the image encoder $\mathcal{E}_{v}$ to retain its feature extraction capability in the training process. 

\textbf{Adapter in image encoder}: Since SAM's image encoder is adapted for purely visual tasks and does not align with text features, directly applying the frozen image encoder for RVOS leads to suboptimal outcomes. To improve the adaptability of the image encoder, we employ an adapter tuning approach \cite{ye2024hi, wu2023medical}. For the latter half of the transformer blocks in the SAM's image encoder, each block is augmented with two trainable adapters. Each adapter consists of a linear layer for dimensionality reduction, a ReLU activation, and a linear layer for dimensionality expansion. We froze the image encoder of SAM and only fine-tuned the Adapter.

\subsubsection{Text Encoder}

At the same time, given the referring expression with $L$ words, we utilize a large language model text-to-text model (T5) \cite{raffel2020exploring} as the text encoder $\mathcal{E}_{t}$ to get the corresponding linguistic embeddings, as shown in the light green part in the bottom of Figure~\ref{fig:overall_network}. Specifically, given the referring expression $E = \{ e_{l} \}_{l=1}^{L}$, these words are firstly tokenized as $T = \{ t_{l} \}_{l=1}^{L}$. Then, we put these tokens into the text encoder to obtain the final embeddings. As our text encoder $\mathcal{E}_{t}$ is reused from the T5 model, we extract the feature vector after the last hidden layer of the encoder in the T5 model as the word embedding, \textit{i.e.},
\begin{equation}
    \mathcal{F}_{e}=\mathcal{E}_{t}(E) \in \mathbb{R}^{L \times C_{e}}.
\end{equation}
Here $\mathcal{F}_{e}$ is a sequence of $C_{e}$-dimensional embeddings of $L$ words, \textit{i.e.}, $\mathcal{F}_{e}=\{f_{i}\}_{i=1}^{L}$, where each word is represented by a $C_{e}$-dimensional embedding. Then the sentence-level embedding $f_{e}^{s} \in \mathbb{R}^{C_{e}}$ can be obtained by simply applying a pooling operation on these word embeddings.  

\subsection{Cross-modal MLP}

Based on the Text Encoder, we get the linguistic embeddings, including word and sentence embeddings. However, there is a significant semantic gap between linguistic embedding space and visual embedding space. Inspired by MiniGPT-4 \cite{zhu2023minigpt}, we design a cross-modal MLP $L_{s}$ which consists of one hidden layer to effectively align the linguistic embedding space and visual embedding space, as shown in the green part in the bottom of Figure~\ref{fig:overall_network}. Specifically, for each word embedding $f_{i}$ in $\mathcal{F}_{e}$, the sparse embedding can be obtained by adopting the cross-modal MLP $L_{s}$, \textit{i.e.}, 
\begin{equation}
    f_{i}^{s} = L_{s}(f_{i}) \in \mathbb{R}^{C_{v}}.
\end{equation} 
In this way, sparse embeddings for $L$ words and the sentence can be obtained, which can be represented as $\mathcal{F}_{sparse}=\{f_{j}^{s}\}_{j=1}^{L}$ and $f_{e^{\prime}}^{s} \in \mathbb{R}^{C_{v}}$, respectively. 

\subsection{Hierarchical Dense Attention}
\begin{figure*}[!t]
\centering
\includegraphics[width=1\linewidth]{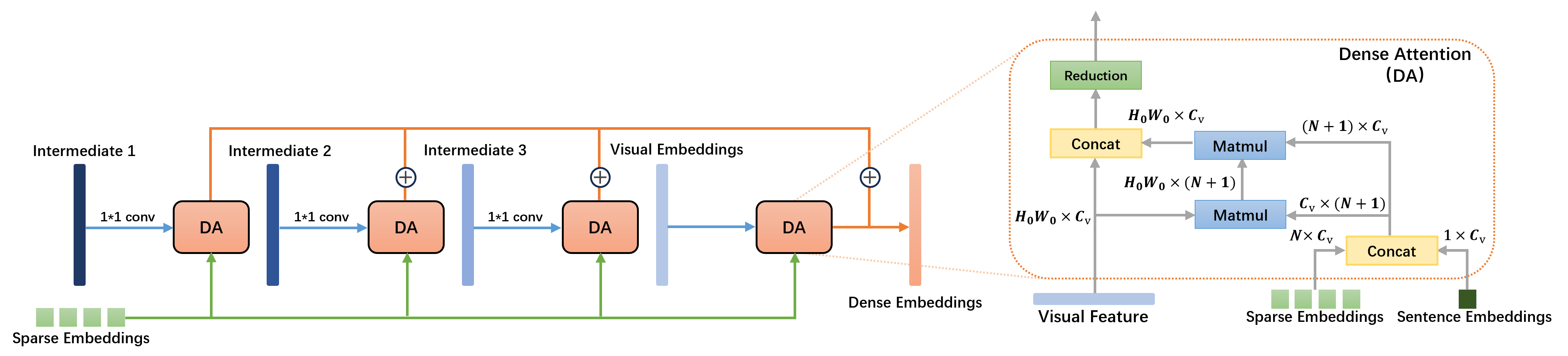}
\caption{The structure of HDA. 1) The left part denotes the overall architecture of HDA. The inputs include the Intermediate Embeddings, Visual Embeddings, and Sparse Embeddings. The output is fine-grained Dense Embeddings. 2) The right part denotes the structure of Dense Attention for fusing hierarchical visual semantic information and sparse embeddings.}
\label{fig:Hierarchical Dense Attention}
\end{figure*}

\subsubsection{Architecture}
In our experiment, we discovered that the original pipeline of SAM, which solely utilizes the output of the Visual Encoder in the mask decoder, struggles to effectively identify tiny objects and leverage intricate text prompts. To address this isuee, we propose a novel HDA module, as shown in Figure~\ref{fig:Hierarchical Dense Attention}, which leverages different levels of visual features and textual features to acquire fine-grain cross-modal embeddings. Specifically, we put intermediate embeddings $ \mathcal{F}_{mid} = \{f^{mid}_{t}\}_{t=1}^{T}$, visual embeddings $ \mathcal{F}_{v} = \{f_{t}\}_{t=1}^{T}$, and sparse embeddings $\mathcal{F}_{sparse}=\{f_{j}^{s}\}_{j=1}^{L}$ together with aligned sentence embeddings $f_{e^{\prime}}^{s}$ into the HDA module. For each feature map, we combine it with sparse embeddings and aligned embeddings in the Dense Attention (DA) module for pixel-level fusion. The output of each Dense Attention will be summed to generate the final Dense Embeddings. For intermediate feature maps, we use an extra $1\times1$ convolutional layer to reduce its dimensionality before feeding them into Dense Attention. 

\subsubsection{Dense Attention}
To enhance the visual features and provide additional cues for the mask decoder, we propose the DA module for fusing visual and linguistic features at the pixel level, as shown in the right part of Figure~\ref{fig:Hierarchical Dense Attention}. For example, given the visual embeddings $ \mathcal{F}_{v}$, we firstly concatenate the aligned sentence embeddings $f_{e^{\prime}}^{s}$ and sparse embeddings $\mathcal{F}_{sparse}=\{f_{j}\}_{j=1}^{L}$ together: 
\begin{equation}
    \mathcal{F}_{fix} =(f_{e^{\prime}}^{s} \copyright \mathcal{F}_{sparse}) \in \mathbb{R}^{(N+1) \times C_{v}}.
\end{equation}
Next, we calculate the similarity between $\mathcal{F}_{fix}$ and $ \mathcal{F}_{v}$ to get the attention map $\mathcal{F}_{sp}$. Then, we apply the dot product between $\mathcal{F}_{sp}$ and $\mathcal{F}_{fix}$ to get the attentive feature, \textit{i.e.}, 
\begin{equation}
    \mathcal{F}_{sl}=\mathcal{F}_{sp} \cdot \mathcal{F}_{fix}.
\end{equation}
Then, we concatenate $\mathcal{F}_{sl}$ and $ \mathcal{F}_{v}$ and apply a convolutional layer to reduce the number of channels to match the channel dimension of $ \mathcal{F}_{v}$, \textit{i.e.},
\begin{equation}
    \mathcal{F}_{dense}=Conv(\mathcal{F}_{sl} \copyright \mathcal{F}_{v}).
\end{equation}
Note that for simplicity we also use $\mathcal{F}_{dense}$ to denote a sequence of dense embeddings for $T$ frames in the video clip, \textit{i.e.}, $ \mathcal{F}_{dense} = \{f_{k}\}_{k=1}^{T}$, where $f_{k} \in \mathbb{R}^{C_{v} \times H_{0} \times W_{0}}$ is the dense embedding for the k$th$ frame. 

\subsection{Mask Decoder}
\label{sec:Mask Decoder}
The Mask Decoder of the vanilla SAM leverages sparse embeddings (point and box) from the prompt encoder and dense embeddings (mask) from the SAM predictor to get final predictions. Following this principle, we construct sparse embeddings and dense embeddings with the same shape which encode useful visual and linguistic features from the Cross-Modal MLP and HDA module. Then we put sparse embeddings $\mathcal{F}_{sparse}$ and dense embeddings $ \mathcal{F}_{dense}$ together with visual embeddings $ \mathcal{F}_{v}$ into the Mask Decoder to get the mask predictions, \textit{i.e.},
\begin{equation}
    M = Decoder(\mathcal{F}_{sparse}, \mathcal{F}_{dense}, \mathcal{F}_{v}),
\end{equation}
where $M$ is the output of the Mask Decoder. Finally, the Mask Decoder uses the output $M$ with the highest score as the final mask prediction.

\subsection{Implicit Tracking Module}
In the context of the RVOS task, leveraging historical cues plays a crucial role. To this end, we devise an ITM $\mathcal{T}$ to transfer historical information to the subsequent frame. Specifically, in each frame, we extract the main mask token $ E_{m} \in \mathbb{R}^{C_{v}}$, which is processed through the Mask Decoder to fuse with visual and text features, and then fed into $\mathcal{T}$ to obtain a track token $E_{track}$. $\mathcal{T}$ consists of two layers of feed-forward neural networks (FFNs) and includes a residual connection, a ReLU activation layer, and a final layer normalization. The shape of $E_{track}$ is same as $ E_{m}$, 
\textit{i.e.},
\begin{equation} 
    E_{track} = \mathcal{T}(E_{m}) \in \mathbb{R}^{C_{v}},
\end{equation}
In current frame, We concatenate all mask tokens and the track token from the last frame in the Mask Decoder. Since $ E_{m}$ encodes the mask of the target object in the current frame, the generated track token $E_{track}$ can provide valuable guidance for the location of the target object in the subsequent frame, \textit{i.e.}, $\mathcal{T}$ implicitly functions as a tracking module.

\subsection{Segmentation with Tracking}
The training and inference processes are illustrated in Figure \ref{fig:track_forward}. For the first frame, the training and inference processes in the Mask Decoder follow the same approach as in Sec. \ref{sec:Mask Decoder}, and we get $E_{m}$ from the Mask Decoder, \textit{i.e.},
\begin{equation}
    M, E_{m} = Decoder(\mathcal{F}_{sparse}, \mathcal{F}_{dense}, \mathcal{F}_{v}),
\end{equation} 
For the subsequent frame, we put sparse embeddings $\mathcal{F}_{sparse}$, dense embeddings $ \mathcal{F}_{dense}$, and visual embeddings $ \mathcal{F}_{v}$ together with the track token of the last frame $E_{track}$ into the Mask Decoder to get the mask prediction and $E_{m}$, \textit{i.e.},
\begin{equation}
    M, E_{m} = Decoder(\mathcal{F}_{sparse}, \mathcal{F}_{dense}, \mathcal{F}_{v}, E_{track}).
\end{equation}  

\section{Experiments}\label{sec4}

\subsection{Experiment Settings} 
\subsubsection{Datasets}
Our experimental evaluations were carried out on two demanding referring video object segmentation datasets: Refer-Youtube-VOS \cite{seo2020urvos} and Ref-DAVIS17 \cite{khoreva2019video}. Refer-Youtube-VOS is an extensive referring video object segmentation dataset that comprises approximately 15,000 referring expressions associated with more than 3,900 videos.
Ref-DAVIS17 is an extension of the DAVIS17 dataset \cite{pont20172017}, where it enhances the dataset by providing language descriptions for each specific object present in the videos. It consists of a total of 90 videos.
We follow the common practice and use the default split for training and testing. 

\subsubsection{Evaluation Metrics}
The main evaluation metrics used in Refer-Youtube-VOS and Ref-DAVIS17 are the average of region similarity ($\mathcal{J}$) and contour accuracy ($\mathcal{F}$), denoted as $\mathcal{J}\&\mathcal{F}$. Regarding Refer-Youtube-VOS, as the annotations for the validation set are not publicly accessible, we assess our method's performance on the official challenge server \footnote{\href{https://codalab.lisn.upsaclay.fr/competitions/3282}{https://codalab.lisn.upsaclay.fr/competitions/3282}}. As for Ref-DAVIS17, it is evaluated using the official evaluation code\footnote{\href{https://github.com/davisvideochallenge/davis2017-evaluation}{https://github.com/davisvideochallenge/davis2017-evaluation}}.

\subsection{Implementation Details}
\subsubsection{Model Details}
We use the image encoder of SAM, \textit{i.e.}, the ViT backbone \cite{dosovitskiy2020image}, as our visual encoder. For the text encoder, we use the Hugging Face \cite{wolf2020transformers} implementation of T5-3b \cite{raffel2020exploring}. The Cross-Modal MLP consists of one hidden layer, which employs the rectified linear unit (ReLU) activation function. In HDA, we utilize the features from three intermediate layers of the visual encoder and the output feature maps of the visual encoder as the input. The convolution layer we use in DA applies a set of $512$ $1 \times 1$ filters to the input feature maps, resulting in 256 output feature maps. ITM consists of two layers of feed-forward neural networks (FFNs) and includes a residual connection, a ReLU activation layer, and a final layer normalization. We freeze the parameters of the Text Encoder during the entire training stage. 
For the Visual Encoder, we tune the added adapter while freezing the other parameters throughout the entire training stage. We add the adapter to the latter half of the transformer blocks in SAM's image encoder. %
We use the output feature maps from the Visual Encoder as the input to Dense Attention Module and Mask Decoder. The dimension of the input embeddings (including visual embeddings, sparse embeddings, and dense embeddings) to the Mask Decoder is $\mathcal{C} = 256$.

\subsubsection{Training and Inference Details}
\label{subsubsec:trainingdetails}
Following \cite{wu2022language, luo2024soc}, our experiments contain the pre-training and fine-tuning process. 
As in SOC\cite{luo2024soc}, we pre-train on the RefCOCO dataset for 40 epochs. 
Then, we fine-tune it on Refer-Youtube-VOS for 4 epochs, and we randomly sample $N=3$ frames in each batch size for training the ITM module. The data augmentation includes random horizontal flip, random resize, random crop, and photometric distortion. The learning rate is set to 1e-4 for the Cross-Modal MLP, HDA, and DA, 1e-6 for the Mask Decoder, and 1e-5 for the Adapter. We employ Dice loss and Focal loss as our primary loss functions. Our model is implemented in PyTorch and trained on 8 NVIDIA Tesla A100 GPUs. During inference, the video frames are down-scaled to 360p. 

\subsection{Main Results}

\begin{table}[!htbp]
\caption{Results on Ref-DAVIS17.} \label{tbl1}
    \centering
    \begin{tabular}{cccc}
    \hline
        Methods & $\mathcal{J}$ \& $\mathcal{F}$ & $\mathcal{J}$ & $\mathcal{F}$ \\ 
    \hline
        CMSA\cite{ye2019cross} & 34.7 & 32.2 & 37.2 \\   
        CMSA+RNN\cite{ye2019cross} & 40.2 & 36.9 & 43.5 \\
        URVOS\cite{seo2020urvos} & 51.5 & 47.3 & 56.0 \\
        ReferFormer\cite{wu2022language} & 61.1 & 58.1 & 64.1 \\
        VLT\cite{ding2021vision} & 61.6 & 58.9 & 64.3 \\
        SgMg\cite{wu2023onlinerefer} & 63.3 & 60.6 & 66.0 \\
        SOC\cite{luo2024soc} & 64.2 & 61.0 & 67.4 \\
        FTVR\cite{li2024fine} & 64.2 & 61.1 & 67.2 \\
        OnlineRefer\cite{miao2023spectrum} & 64.8 & 61.6 & 67.7 \\
    \hline
        PerSAM\cite{zhang2023personalize}+Grounding Dino & 40.3 & 38.2 & 42.4 \\
        SAM-Track\cite{cheng2023segment}+Grounding Dino & 57.1 & 54.6 & 59.7 \\
    \hline
        \textbf{RefSAM} & \textbf{71.9} & \textbf{68.4} & \textbf{75.5} \\
    \hline
    \end{tabular}
\end{table}

\subsubsection{Ref-DAVIS17}

We conduct experiments on the Ref-DAVIS17 dataset, and the results are shown in Table \ref{tbl1}. We can find that our method RefSAM can achieve 71.9 in terms of $\mathcal{J}$ \& $\mathcal{F}$, which is 7.1 points higher than the state-of-the-art method OnlineRefer \cite{miao2023spectrum}, which verifies the superiority of our method. It is attributed to the excellent capability of object segmentation and cross-modal understanding. Consequently, by taking advantage of the powerful foundation model SAM, our RefSAM can correctly identify and accurately segment the target object in the video clip. 

To have a closer look at the effectiveness of our method, we also compare our method with some recent SAM-based methods. Since there are no off-the-shelf SAM-based models for RVOS, existing SAM-based methods, such as SAM-Track \cite{cheng2023segment} and PerSAM \cite{zhang2023personalize}, can be adapted to RVOS task by providing bounding boxes in the first frame. Specifically, simple SAM-based RVOS baselines can be derived by combining SAM-Track or PerSAM with Grounding Dino \cite{liu2023grounding}, which is an object detector that can provide bounding boxes of objects in the first frame with referring expressions. As shown in Table \ref{tbl1}, these naive solutions exhibit satisfactory performance, surpassing the CMSA and URVOS methods, yet significantly lagging behind our RefSAM model. Firstly, these methods employ a two-stage pipeline that can lead to a sub-optimal performance in the downstream segmentation task due to inaccurate bounding boxes detected by the object detector in the initial frame. Secondly, the absence of specific fine-tuning on the RVOS dataset poses a challenge, primarily due to the substantial model size and the inherent two-stage design. It is also noteworthy that the involvement of individual models for each sub-task also brings extra difficulties for model deployment. These findings underscore the efficacy of our RefSAM's end-to-end design and parameter-efficient tuning strategy.

Finally, to demonstrate that fine-tuning SAM is superior to directly using SAM for post-processing, we apply SAM to the post-processing of FTVR's inference results. For each inference result from FTVR, we extract the result's bounding box and input it into SAM for more fine-grained segmentation. 
Using SAM for post-processing the results of FTVR improves the J\&F score by 2.5 points, raising it to 66.7. However, this score is still 5.2 points lower than RefSAM, demonstrating the efficacy of fine-tuning SAM.

\subsubsection{Refer-Youtube-VOS}

\begin{table}[!ht]
\caption{Results on Refer-Youtube-VOS.} \label{youtube_table}
    \centering
    \begin{tabular}{cccc}
    \hline
        Methods &  $\mathcal{J}$ \& $\mathcal{F}$ & $\mathcal{J}$ & $\mathcal{F}$ \\ 
    \hline
        CMSA \cite{ye2019cross} & 34.9 & 33.3 & 36.5 \\
        CMSA + RNN \cite{ye2019cross} & 36.4 & 34.8 & 38.1 \\
        URVOS \cite{seo2020urvos} & 47.2 & 45.3 & 49.2 \\
        PMINet \cite{ding2021progressive} & 48.2 & 46.7 & 49.6 \\
        PMINet + CFBI \cite{ding2021progressive} & 54.2 & 53.0 & 55.5 \\
        MTTR \cite{botach2022end} & 55.3 & 54.0 & 56.6 \\
        CITD \cite{liang2021rethinking} & 61.4 & 60.0 & 62.7 \\
        FTVR \cite{li2024fine} & 62.7 & 60.9 & 64.5 \\
        ReferFormer \cite{wu2022language} & 62.9 & 61.3 & 63.5 \\
        OnlineRefer \cite{miao2023spectrum} & 63.5 & 61.6 & 65.5 \\        
        VLT \cite{ding2021vision} & 63.8 & 61.9 & 65.6 \\
        SgMg \cite{wu2023onlinerefer} & 65.7 & 63.9 & 67.4 \\
        SOC \cite{luo2024soc} & 66.0 & 64.1 & 67.9 \\
    \hline
        PerSAM \cite{zhang2023personalize}+Grounding Dino & 30.8 & 27.1 & 34.4 \\
        SAM-Track\cite{cheng2023segment}+Grounding Dino & 52.2 & 49.9 & 54.5 \\
    \hline
        \textbf{RefSAM} & \textbf{67.6} & \textbf{65.8} & \textbf{69.4} \\ 
    \hline
    \end{tabular}
\end{table}

We also conduct experiments on the Refer-Youtube-VOS dataset, and the results are summarized in Table~\ref{youtube_table}. Our RefSAM model achieves a notable performance of 67.6 in terms of $\mathcal{J}$ \& $\mathcal{F}$, surpassing the PerSAM and SAM-Track baselines. Additionally, it surpasses previous methods like VLT, SgMg, and SOC, achieving a 1.6-point improvement over the state-of-the-art SOC. 
In the future, we plan to improve our model by investigating more advanced designs such as more effective cross-modal fusion. 

\subsection{Ablation Study}

We conduct detailed ablation studies on the proposed RefSAM using ViT-B as the backbone, analyzing the impact of the key modules and the influence of hyper-parameters.

\subsubsection{Effect of the Key Modules}

\begin{table*}[!ht]
\caption{Ablation study of the key components of RefSAM.} \label{tbl2}
    \centering 
    \begin{tabular}{ccccccccc}
        \hline
            \multirow{2}{*}{Cross-modal MLP} & \multirow{2}{*}{DA} & \multirow{2}{*}{HDA} & \multicolumn{3}{c}{Refer-Youtube-VOS} & \multicolumn{3}{c}{Ref-DAVIS17} \\
            \cmidrule(lr){4-6} \cmidrule(lr){7-9} 
            &  &  & $\mathcal{J}$ \& $\mathcal{F}$ & $\mathcal{J}$ & $\mathcal{F}$ & $\mathcal{J}$ \& $\mathcal{F}$ & $\mathcal{J}$ & $\mathcal{F}$ \\        
        \hline
            \checkmark &            &            & 48.5 & 47.8 & 49.1 & 57.1 & 53.9 & 60.3  \\
                       & \checkmark &            & 49.7 & 49.0 & 50.3 & 57.8 & 54.8 & 60.9  \\
                       &            & \checkmark & 51.8 & 51.9 & 51.8 & 59.1 & 56.2 & 61.9  \\
            \checkmark & \checkmark &            & 50.2 & 49.8 & 50.6 & 59.3 & 56.6 & 62.1  \\
            \checkmark &            & \checkmark & \textbf{52.2} & \textbf{52.0} & \textbf{52.5} & \textbf{60.3} & \textbf{57.3} & \textbf{63.3}  \\
        \hline
    \end{tabular}
\end{table*}

RefSAM employs Cross-Modal MLP, HDA, and ITM for referring video object segmentation. Since ITM requires sampling $N$ frames for each batch size, which differs from the other modules in the experiment, we first investigate the effects of other key components by isolating ITM. Table \ref{tbl2} summarizes the results on Refer-Youtube-VOS and Ref-DAVIS17. DA means we only use the visual embeddings for Dense Attention. HDA means we use both visual embeddings and intermediate embeddings for Hierarchical Dense Attention. We use the data augmentation described in Sec.~\ref{subsubsec:trainingdetails}. As shown in the example of Ref-DAVIS17, using both DA and Cross-modal MLP instead of HDA and Cross-modal MLP results in a performance drop of 1.0 in the J\&F-Mean score. Additionally, using only the Cross-modal MLP or DA results in performance drops of 3.2 and 2.5 in the J\&F-Mean score, respectively, compared to the best choice. On the other hand, using only HDA increases the performance by 1.3 compared to using only DA, but it still underperforms compared to using both HDA and Cross-modal MLP. 
On Refer-Youtube-VOS, using only HDA outperforms both Cross-modal MLP and DA. 
These results validate the importance of Cross-Modal MLP and HDA in aligning and fusing visual and language features. 

\begin{table}[!ht]
    \centering
    \caption{Ablation study of ITM.}\label{ablation_memory_method}
    \begin{tabular}{cccccc}
        \hline
            Setting & Resid. & $\mathcal{J}$ \& $\mathcal{F}$ & $\mathcal{J}$ & $\mathcal{F}$\\
        \hline
            Sing frame (Tab.~\ref{tbl2})  & & 60.3 & 57.3 & 63.3 \\
            Three frames  & & 59.6 & 56.5 & 62.7 \\
        \hline
            IoU token  & & 61.6 & 58.8 & 64.4 \\
            All mask tokens  & & 61.9 & 58.9 & 64.8 \\
            Main mask token  & & 62.2 & 58.9 & 65.5 \\
            Main mask token & \checkmark & \textbf{62.5} & \textbf{59.5} & \textbf{65.4} \\
        \hline
    \end{tabular}
\end{table}

Next, we ablate the design choice of ITM. The mask decoder of SAM outputs an IoU token, a main mask token, and three-scale mask tokens. We chose the IoU token, main mask token, and all four mask tokens separately as the input of ITM. For comparison, we conducted experiments with two control groups without ITM: one using a single frame and another using three frames as input. For all experiments with ITM, we utilize three frames in each batch size, employ Cross-modal MLP and HDA, and use the data augmentation described in Sec.~\ref{subsubsec:trainingdetails}. The results on Ref-DAVIS17 are shown in Table~\ref{ablation_memory_method}. As can be seen, relying solely on three frames without incorporating other tracking or memory modules can decrease performance. This may be due to the model learning specialized knowledge from different frames instead of general knowledge, thus reducing the model's generalization ability. 
Then, utilizing the main mask token as the input of ITM achieves the best performance with a J\&F-Mean of 62.2, surpassing both the IoU token and all mask tokens. It is reasonable since the main mask token has a better capability of understanding general objects and can provide more accurate location information for a complete object than the other token that either emphasizes different granularities of objects or is responsible for predicting the IoU score.
Furthermore, we investigate the structure design of ITM, \textit{i.e.}, with or without using the residual connection. The results in the last two rows show that utilizing a residual connection in ITM leads to an increase of 0.3 J\&F-Mean.

\subsubsection{Influence of Hyper-parameters}

\begin{figure*}
    \centering
    
    \begin{subfigure}{0.32\textwidth}
        \centering
        \includegraphics[width=\linewidth]{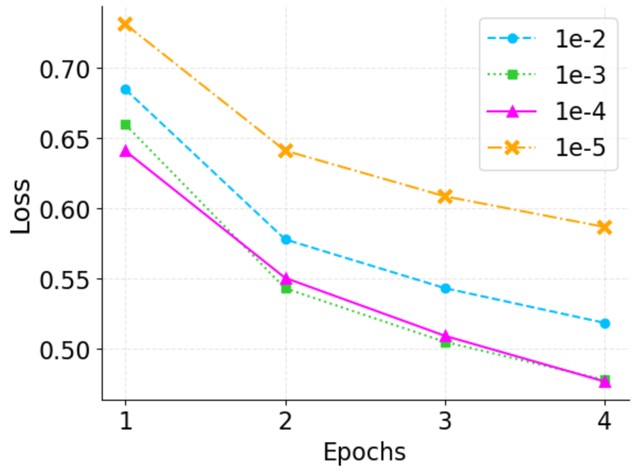}
        \caption{Cross-Modal MLP}
        \label{fig:Learning Rate of Cross Modal MLP}
    \end{subfigure}
    \hfill
    \begin{subfigure}{0.32\textwidth}
        \centering
        \includegraphics[width=\linewidth]{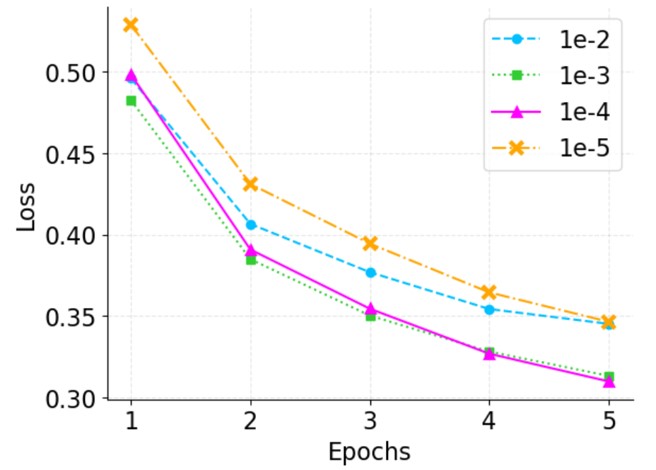}
        \caption{Dense Attention}
        \label{fig:Learning Rate of Dense Attention Conv}
    \end{subfigure}
    \hfill
    \begin{subfigure}{0.32\textwidth}
        \centering
        \includegraphics[width=\linewidth]{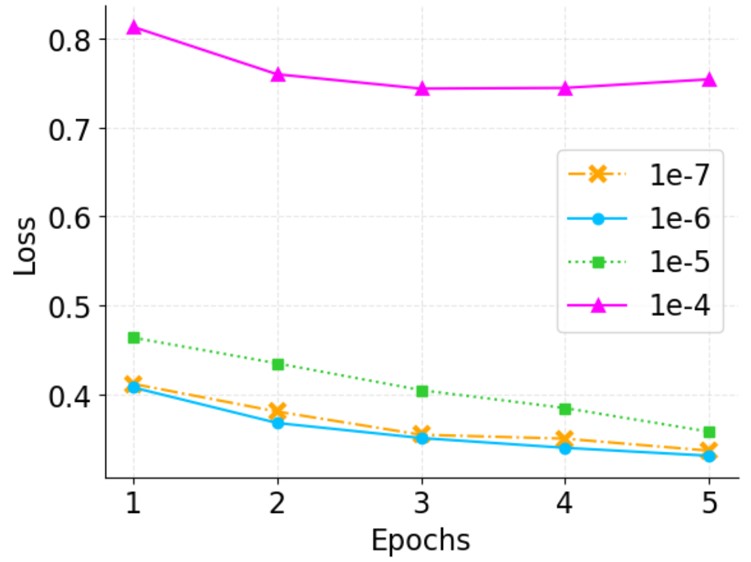}
        \caption{Mask Decoder}
        \label{fig:Learning Rate of Mask Decoder}
    \end{subfigure}
    
    \caption{The influence of different learning rates for the learnable modules of RefSAM.}
    \label{fig:learningrate}
\end{figure*}

\begin{figure}
    \centering
    \includegraphics[width=0.7\linewidth]{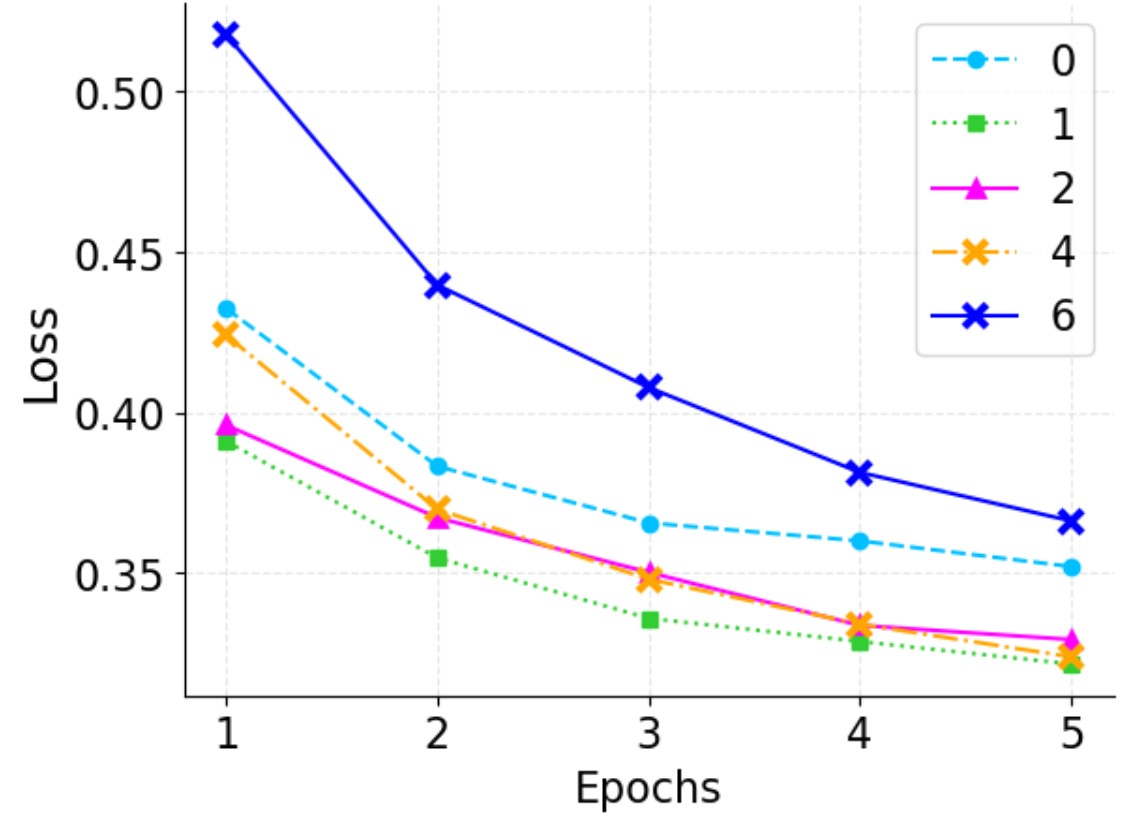}
    \caption{The influence of the number of hidden layers in Cross-Modal MLP.}
    \label{fig:The Number of Hidden Layer in Cross Modal MLP}
\end{figure}

Figure \ref{fig:learningrate} presents the results of different learning rate settings for the learnable modules of RefSAM, including the Cross-Modal MLP, DA, and the Mask Decoder. As can be seen, RefSAM converges faster and better when the learning rate is set to 1e-4 for the Cross-Modal MLP and DA, and 1e-6 for the Mask Decoder. Furthermore, we study the influence of the number of hidden layers in the Cross-Model MLP. As shown in Figure~\ref{fig:The Number of Hidden Layer in Cross Modal MLP}, using a single hidden layer delivers faster and better convergence, which is the default setting.

\subsubsection{Impact of Different Text Encoders}

We also study the impact of different text encoders in RefSAM, including 1) T5-3b \cite{raffel2020exploring}, Roberta-base \cite{liu2019roberta}, and Clip-large~\cite{radford2021learning}. All experiments use the data augmentation described in Sec.~\ref{subsubsec:trainingdetails} and the model employs Cross-modal MLP and HDA. 
The results on Ref-DAVIS17 are shown in \ref{diff_text_encoder}. As can be seen, RefSAM with the T5-3b text encoder performs better than the others, yielding an improvement of 1.6 J\&F-Mean compared to Roberta-base and a more significant increase of 2.0 J\&F-Mean over Clip-large.

\begin{table}[!ht]
\caption{Ablation study of the Text Encoder.} \label{diff_text_encoder}
    \centering
    \begin{tabular}{cccc}
    \hline
        Text Encoder & $\mathcal{J}$ \& $\mathcal{F}$ & $\mathcal{J}$ & $\mathcal{F}$ \\ 
    \hline
        T5-3b \cite{raffel2020exploring} (Tab.~\ref{tbl2}) & \textbf{60.3} & \textbf{57.3} & \textbf{63.3} \\
        Roberta-base \cite{liu2019roberta} & 58.7 & 55.6 & 61.7 \\
        Clip-large \cite{radford2021learning} & 58.3 & 56.1 & 61.3 \\
    \hline
    \end{tabular}
\end{table}

\subsubsection{Influence of Adapter}
In this section, we examine the impact of the Adapter. All experiments use the data augmentation described in Sec.~\ref{subsubsec:trainingdetails} and the model employs Cross-modal MLP and HDA. The results on Refer-Youtube-VOS and Ref-DAVIS17 are shown in \ref{no/adapter}. We conducted experiments on ViT-B. As shown, using ITM or the Adapter improves performance on both Refer-Youtube-VOS and Ref-DAVIS17. Moreover, using both ITM and the Adapter together results in even better performance. 
Using ViT-B on Refer-Youtube-VOS as an example, not using ITM or the Adapter yields a J\&F-Mean score of 52.2 points. Adding ITM or the Adapter increases the score by 2.5 or 4.0 points, respectively. Finally, using both ITM and the Adapter together achieves a J\&F-Mean score of 58.4 points, an increase of 6.2 points.

\begin{table*}[!ht]
\caption{Ablation study of adapter.} \label{no/adapter}
    \centering 
    \begin{tabular}{cccccccc}
        \hline
            \multirow{2}{*}{ITM} & \multirow{2}{*}{Adapter} & \multicolumn{3}{c}{Refer-Youtube-VOS} & \multicolumn{3}{c}{Ref-DAVIS17} \\
            \cmidrule(lr){3-5} \cmidrule(lr){6-8} 
            &  & $\mathcal{J}$ \& $\mathcal{F}$ & $\mathcal{J}$ & $\mathcal{F}$ & $\mathcal{J}$ \& $\mathcal{F}$ & $\mathcal{J}$ & $\mathcal{F}$ \\ 
        \hline
                       &            & 52.2 & 52.0 & 52.5 & 60.3 & 57.3 & 63.3  \\
            \checkmark &            & 54.7 & 53.9 & 55.5 & \textbf{62.5} & \textbf{59.5} & \textbf{65.4}  \\
                       & \checkmark & 56.2 & 55.3 & 57.1 & 61.9 & 58.5 & 65.5  \\
            \checkmark & \checkmark & \textbf{58.4} & \textbf{57.4} & \textbf{59.4} & 62.1 & 59.0 & 65.3  \\
        \hline
    \end{tabular}
\end{table*}

\begin{figure*}[!ht]
    \centering
    \hspace{1cm}
    \begin{subfigure}{\textwidth}
        \centering
        \includegraphics[width=0.22\linewidth]{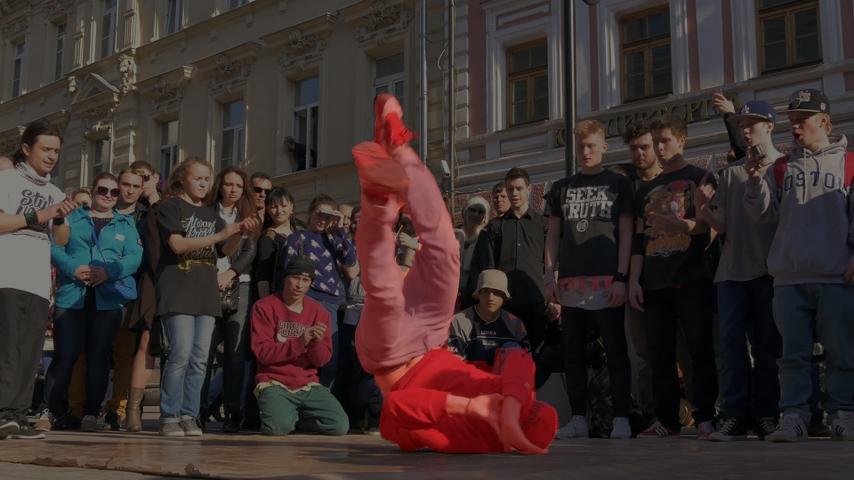}
        \includegraphics[width=0.22\linewidth]{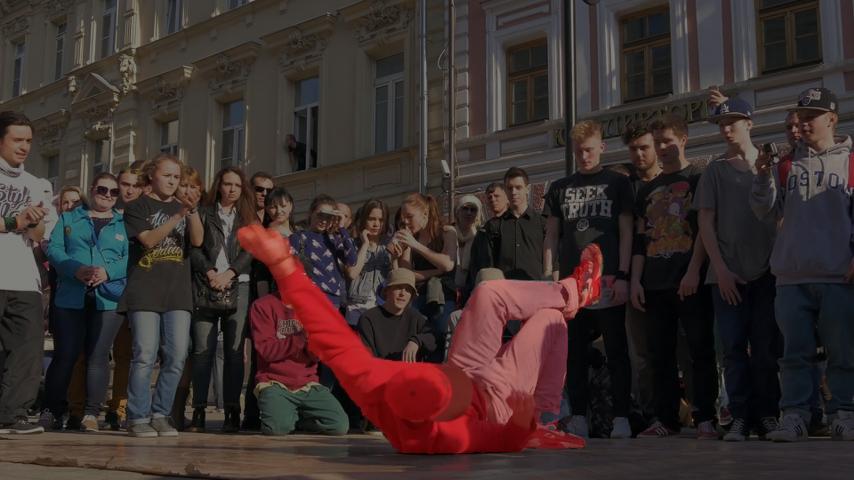}
        \includegraphics[width=0.22\linewidth]{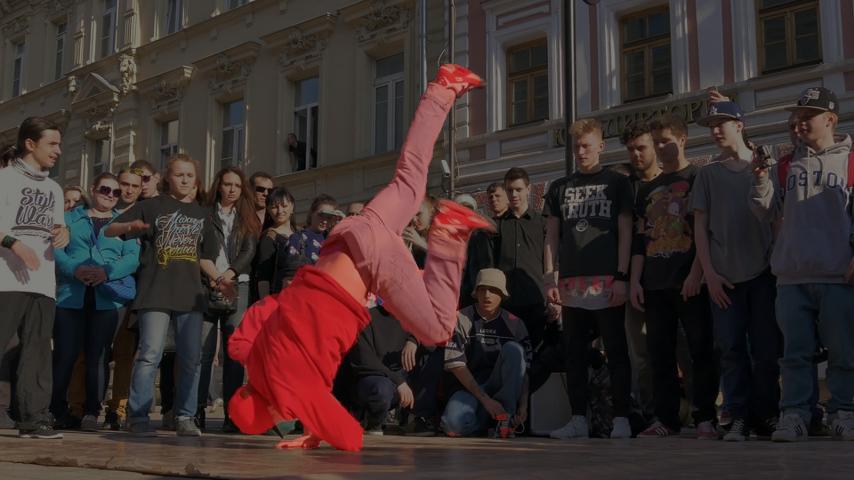}
        \includegraphics[width=0.22\linewidth]{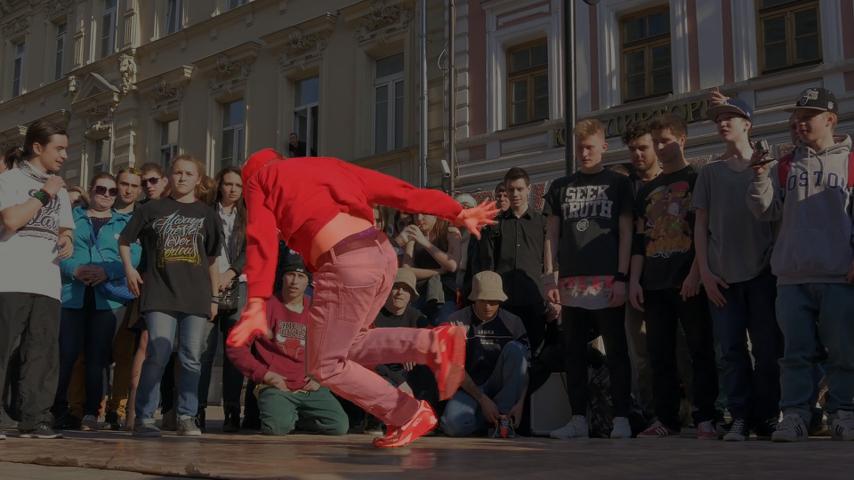}
        \caption{\textcolor{red}{A man in a red sweatshirt performing breakdance}.}
        \label{fig:multiple_images1}
    \end{subfigure}

    \hspace{1cm}
    \begin{subfigure}{\textwidth}
        \centering
        \includegraphics[width=0.22\linewidth]{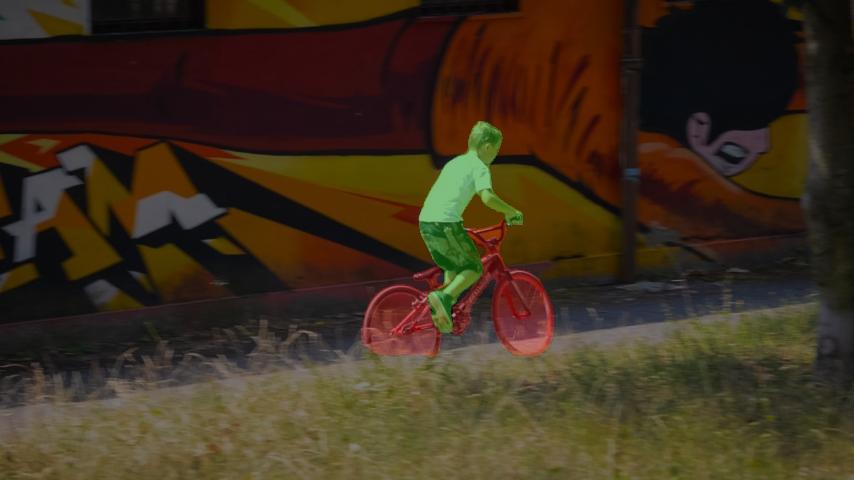}
        \includegraphics[width=0.22\linewidth]{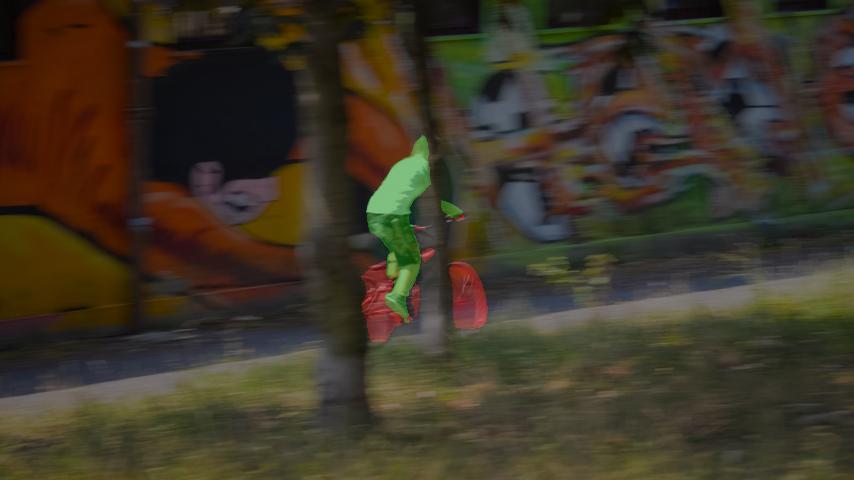}
        \includegraphics[width=0.22\linewidth]{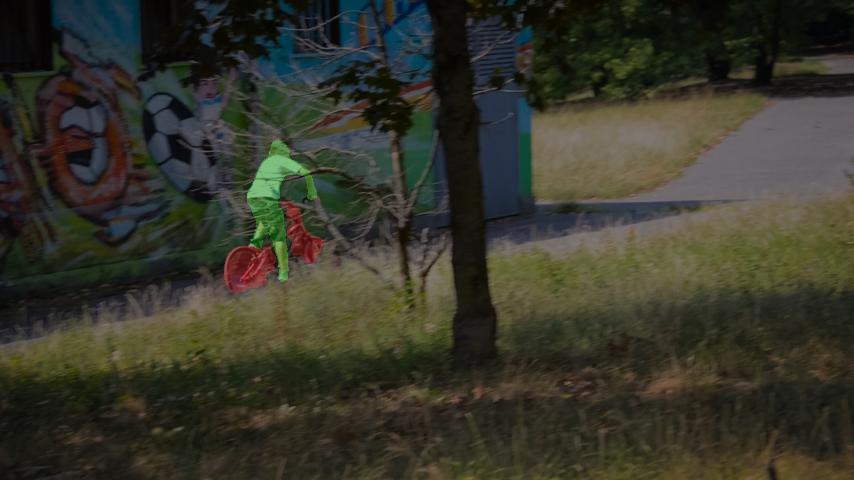}
        \includegraphics[width=0.22\linewidth]{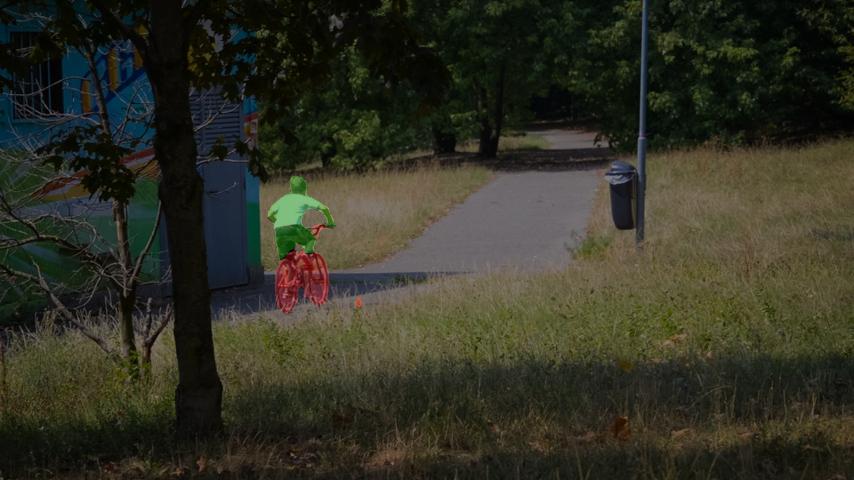}
        \caption{\textcolor{red}{A red bmx bike}. \textcolor{green}{A boy wearing a white tshirt}.}
        \label{fig:multiple_images2}
    \end{subfigure}

    \hspace{1cm}
    \begin{subfigure}{\textwidth}
        \centering
        \includegraphics[width=0.22\linewidth]{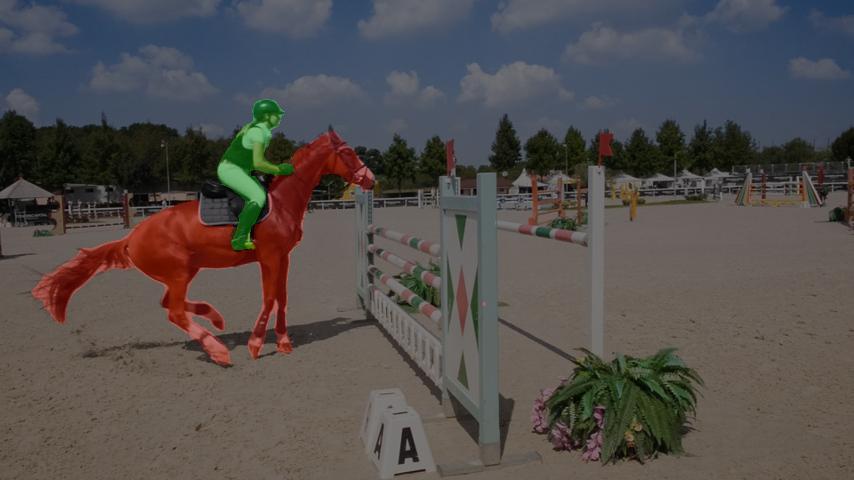}
        \includegraphics[width=0.22\linewidth]{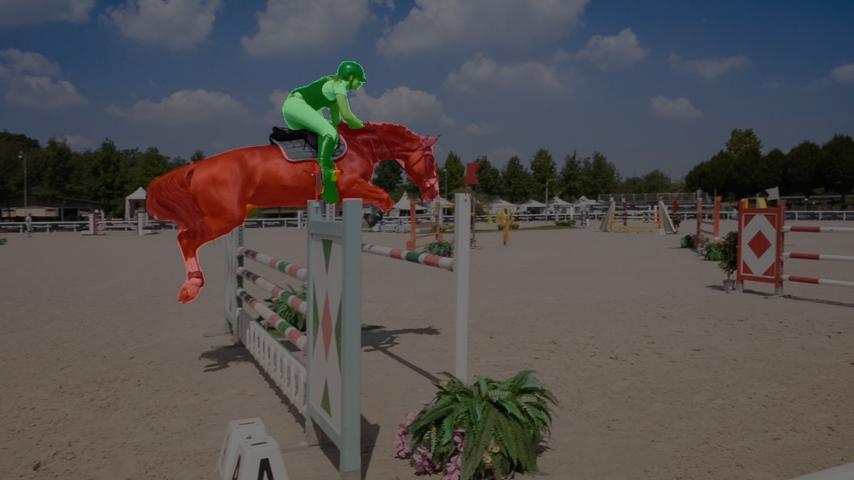}
        \includegraphics[width=0.22\linewidth]{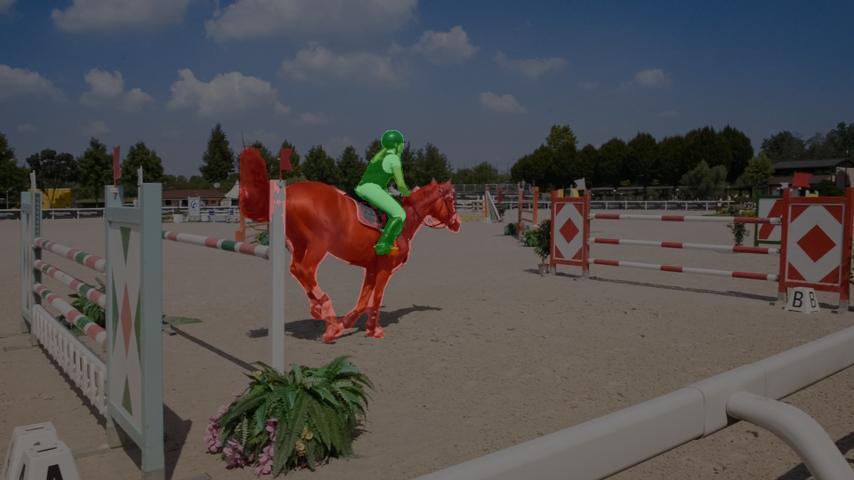}
        \includegraphics[width=0.22\linewidth]{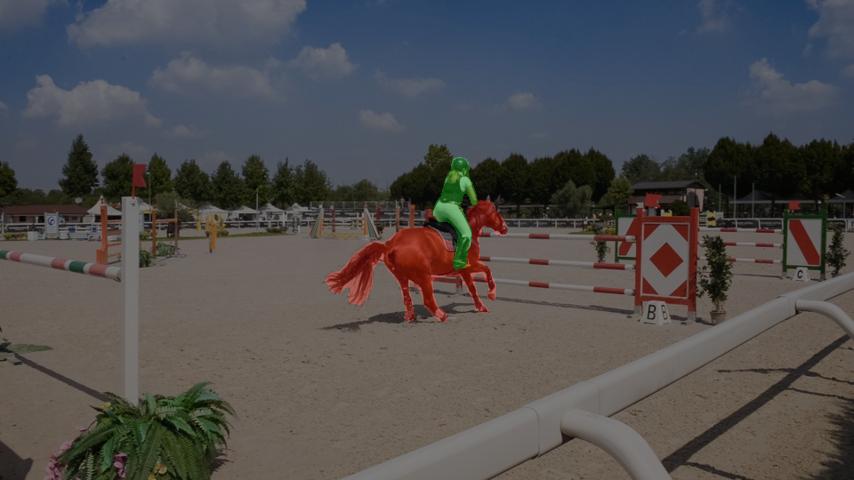}
        \caption{\textcolor{red}{A horse doing high jumps}. \textcolor{green}{A woman riding a horse}.}
        \label{fig:multiple_images3}
    \end{subfigure}
    
    \caption{Visualization of RefSAM's results on Ref-DAVIS17. The text with a specific color corresponds to the object mask of the same color.}
    \label{fig:visualresultsRefSAM}
\end{figure*}

\begin{figure*}[!ht]
    \centering

    \hspace{1cm}
    \begin{subfigure}{\textwidth}
        \centering
        \includegraphics[width=0.22\linewidth]{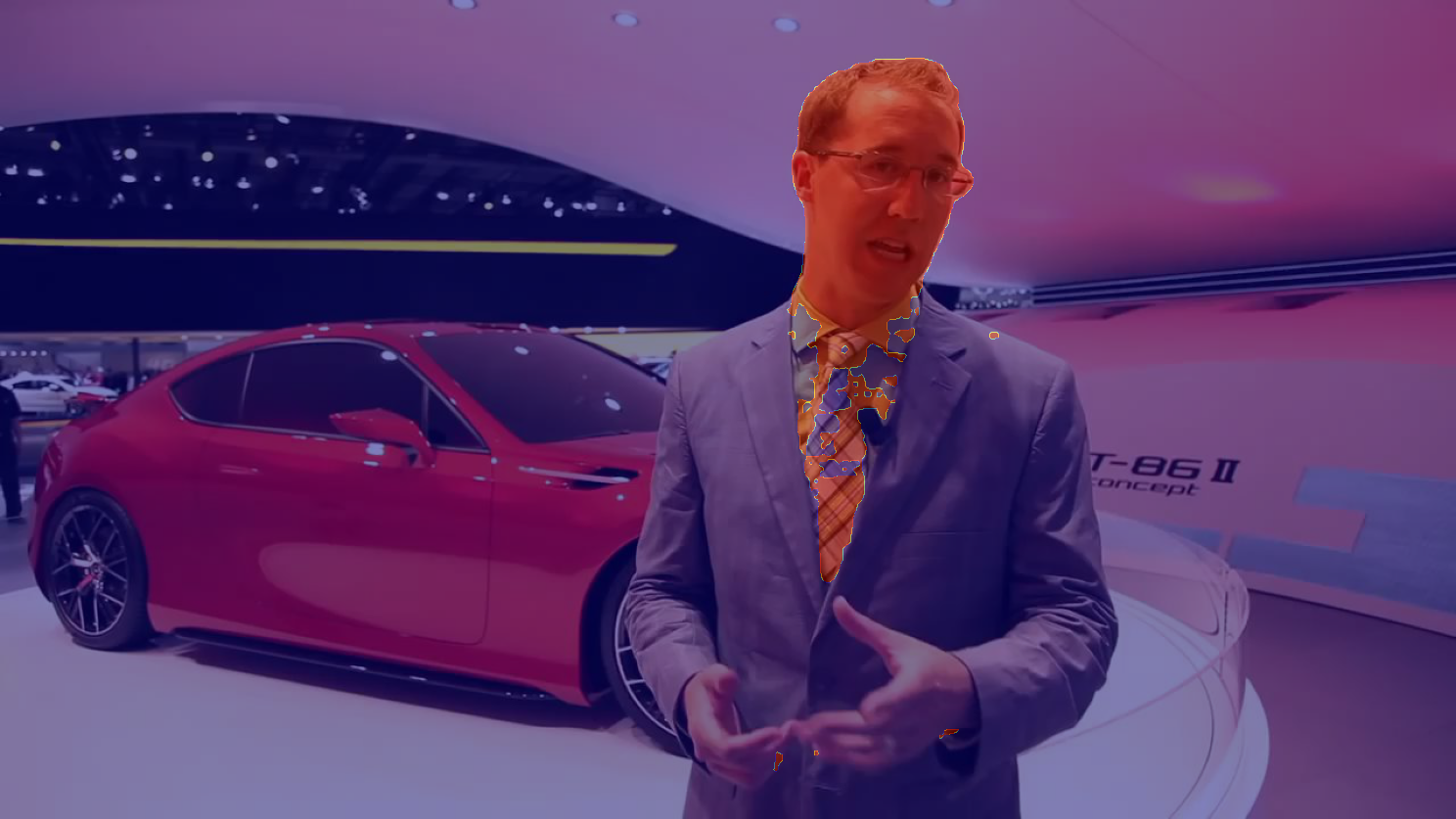}
        \includegraphics[width=0.22\linewidth]{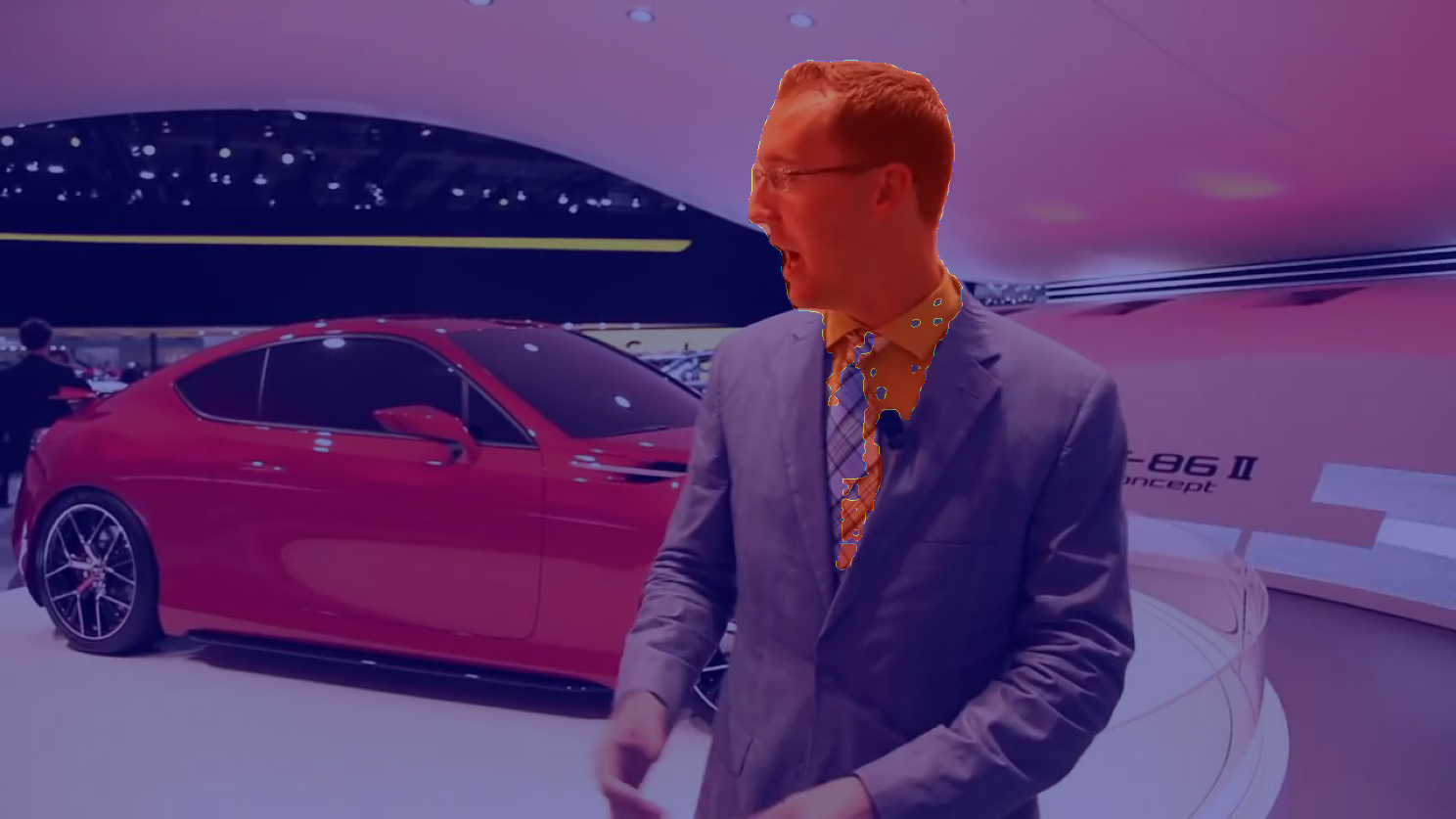}
        \includegraphics[width=0.22\linewidth]{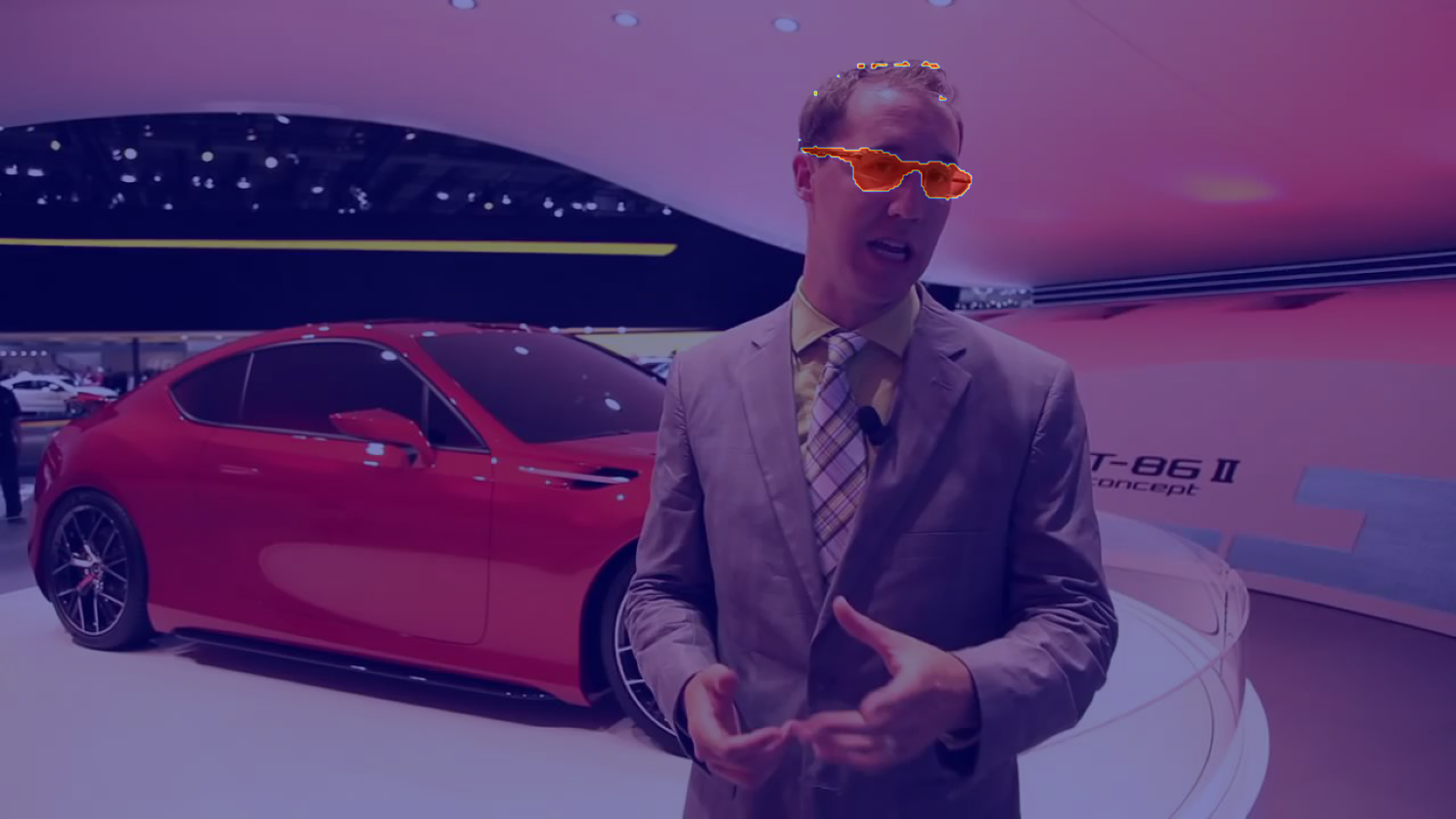}
        \includegraphics[width=0.22\linewidth]{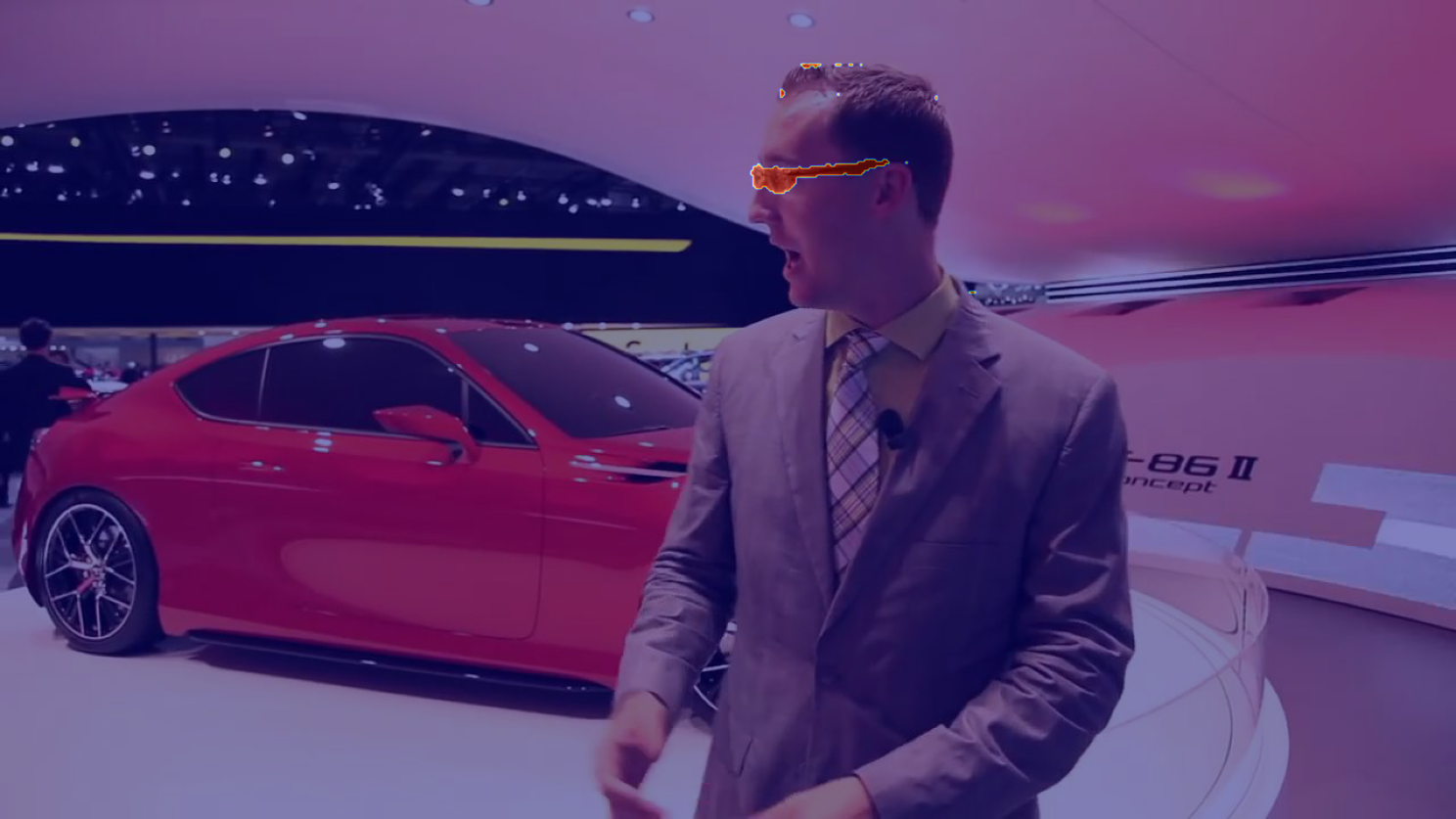}
        \caption{\textcolor{red}{the glasses worn by a man standing in front of a red display car}.}
        \label{fig:multiple_images51}
    \end{subfigure}

    \hspace{1cm}
    \begin{subfigure}{\textwidth}
        \centering
        \includegraphics[width=0.22\linewidth]{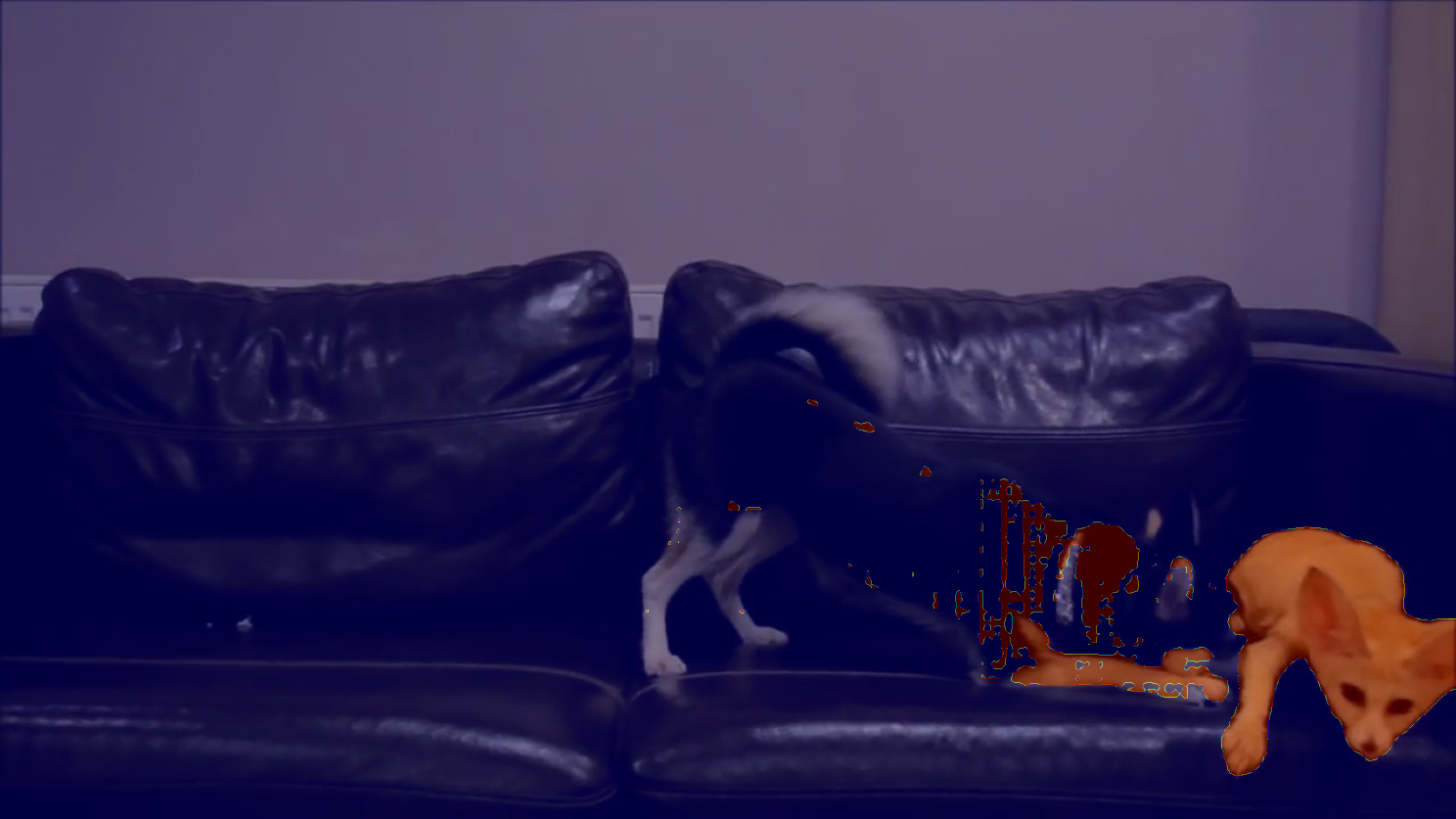}
        \includegraphics[width=0.22\linewidth]{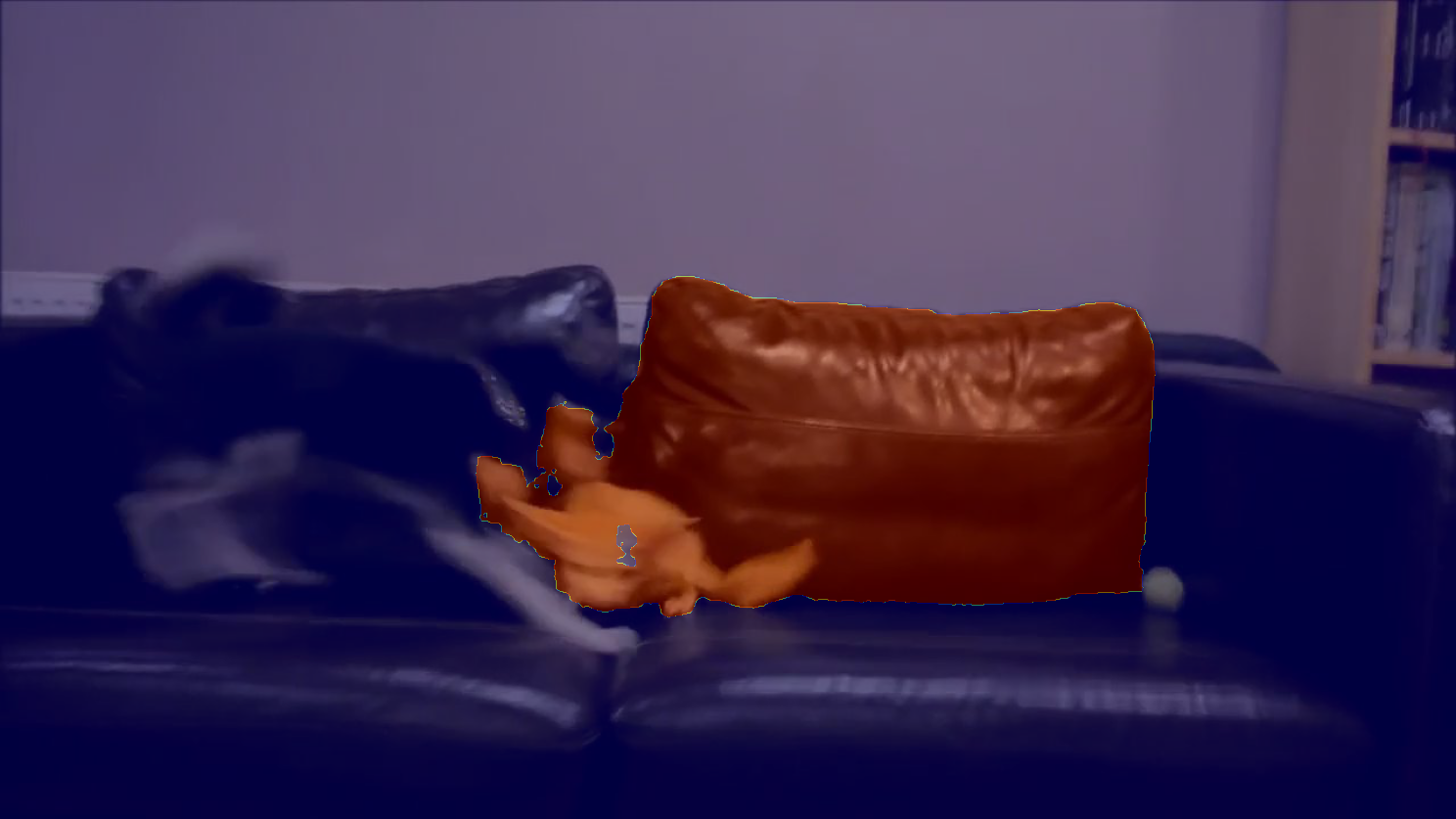}
        \includegraphics[width=0.22\linewidth]{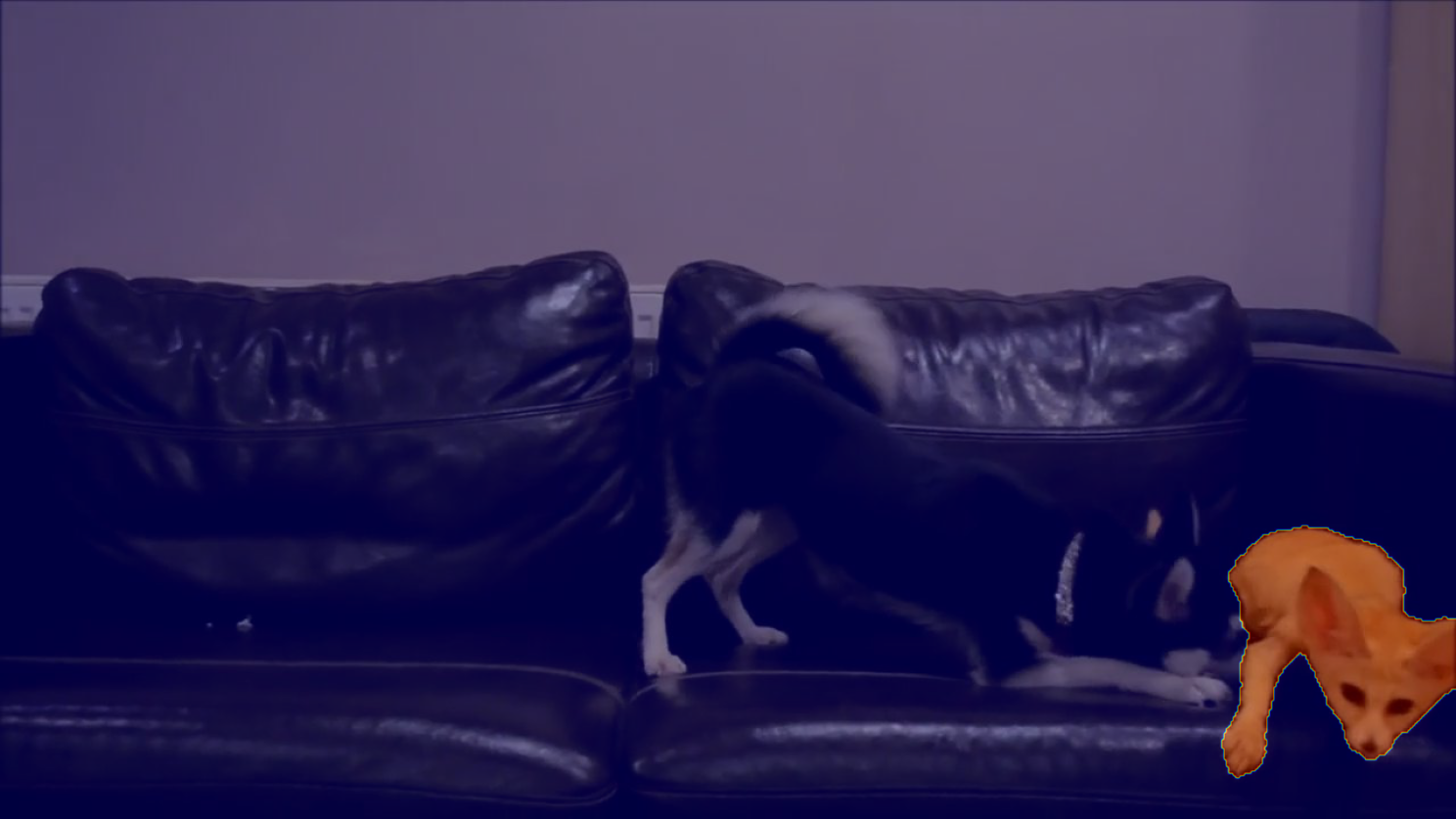}
        \includegraphics[width=0.22\linewidth]{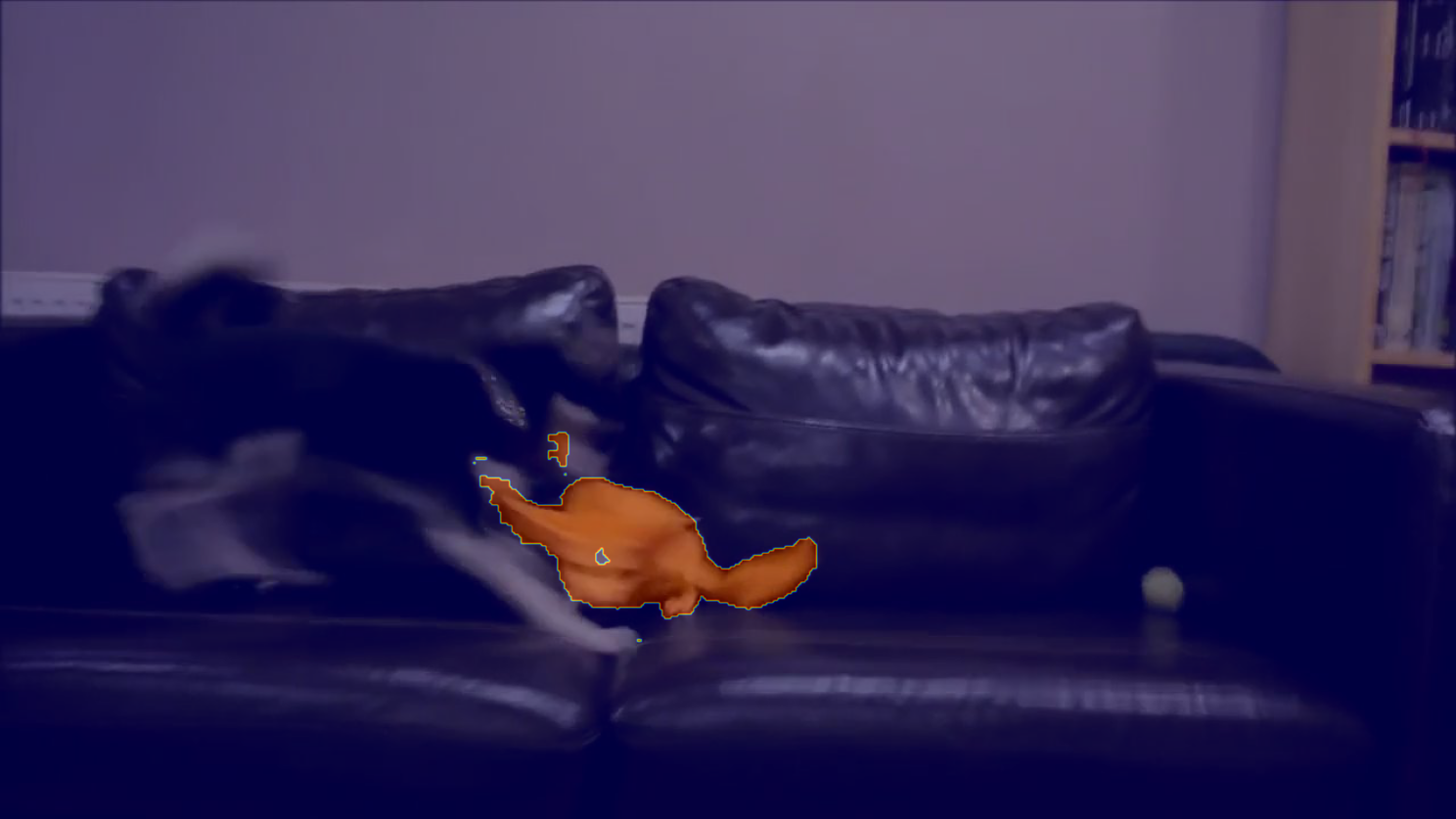}
        \caption[width=0.4\linewidth]{\textcolor{red}{a small fox like dog on the right side of a sofa}.}
        \label{fig:multiple_images52}
    \end{subfigure}

    \hspace{1cm}
    \begin{subfigure}{\textwidth}
        \centering
        \includegraphics[width=0.22\linewidth]{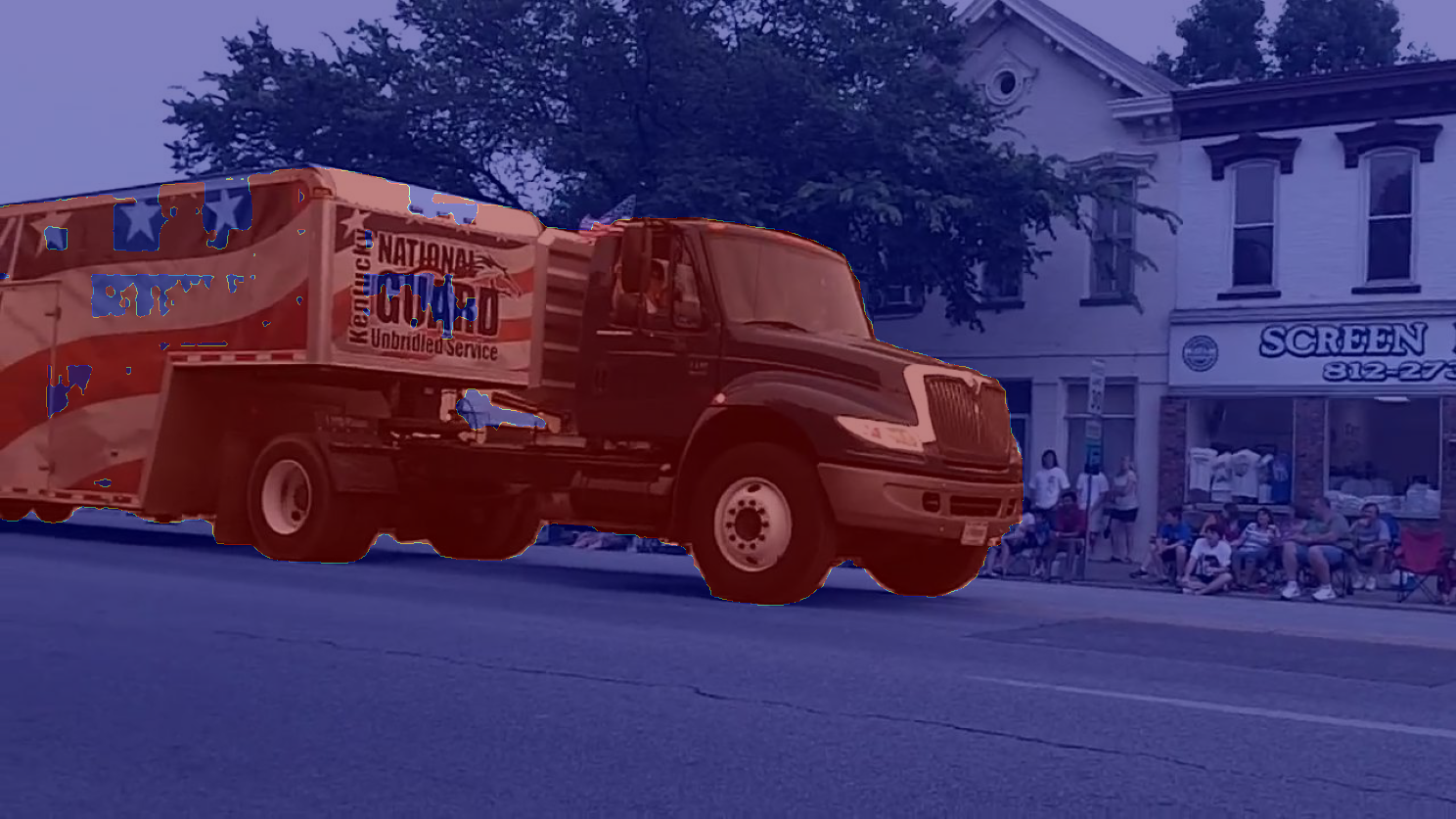}
        \includegraphics[width=0.22\linewidth]{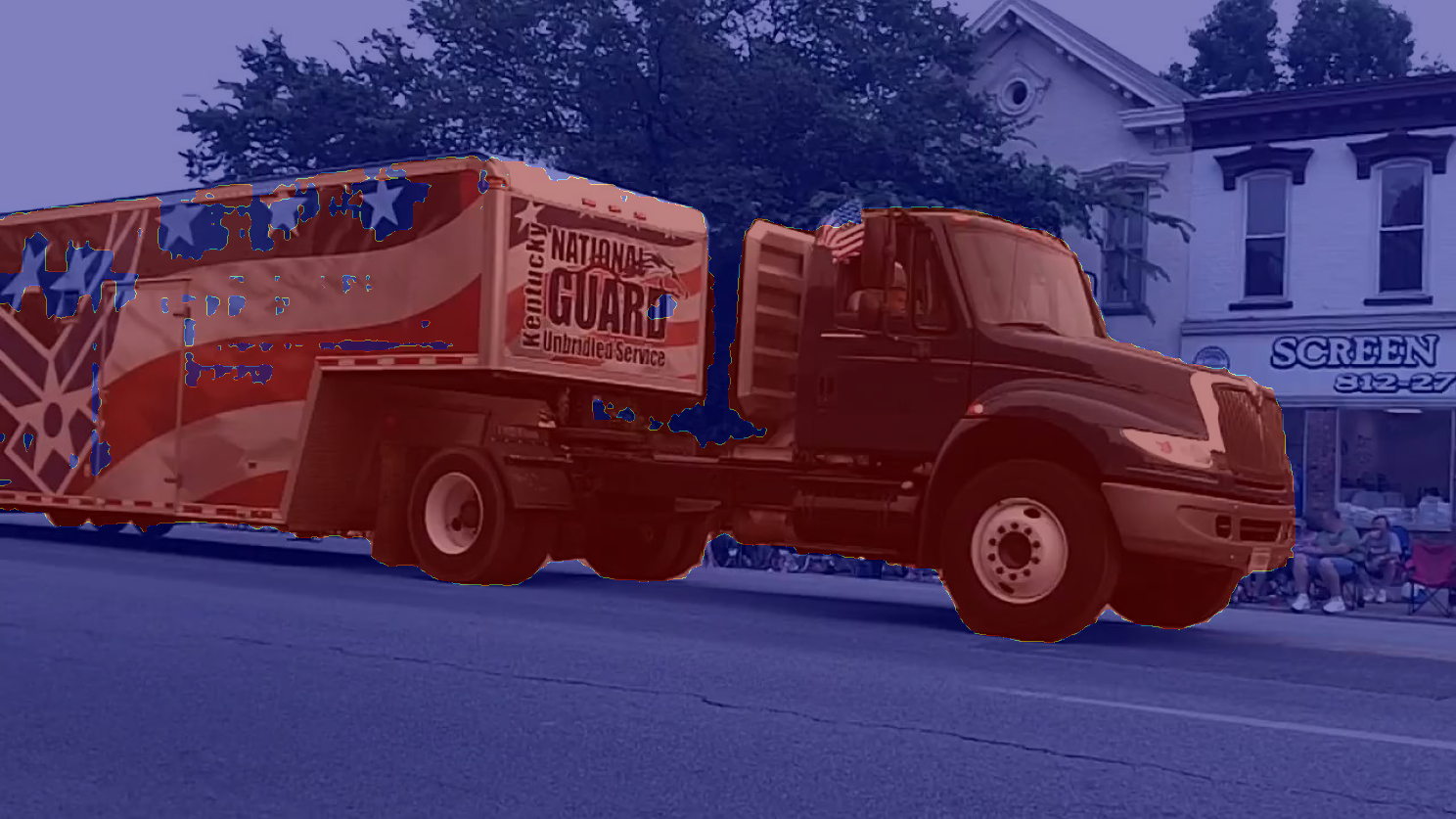}
        \includegraphics[width=0.22\linewidth]{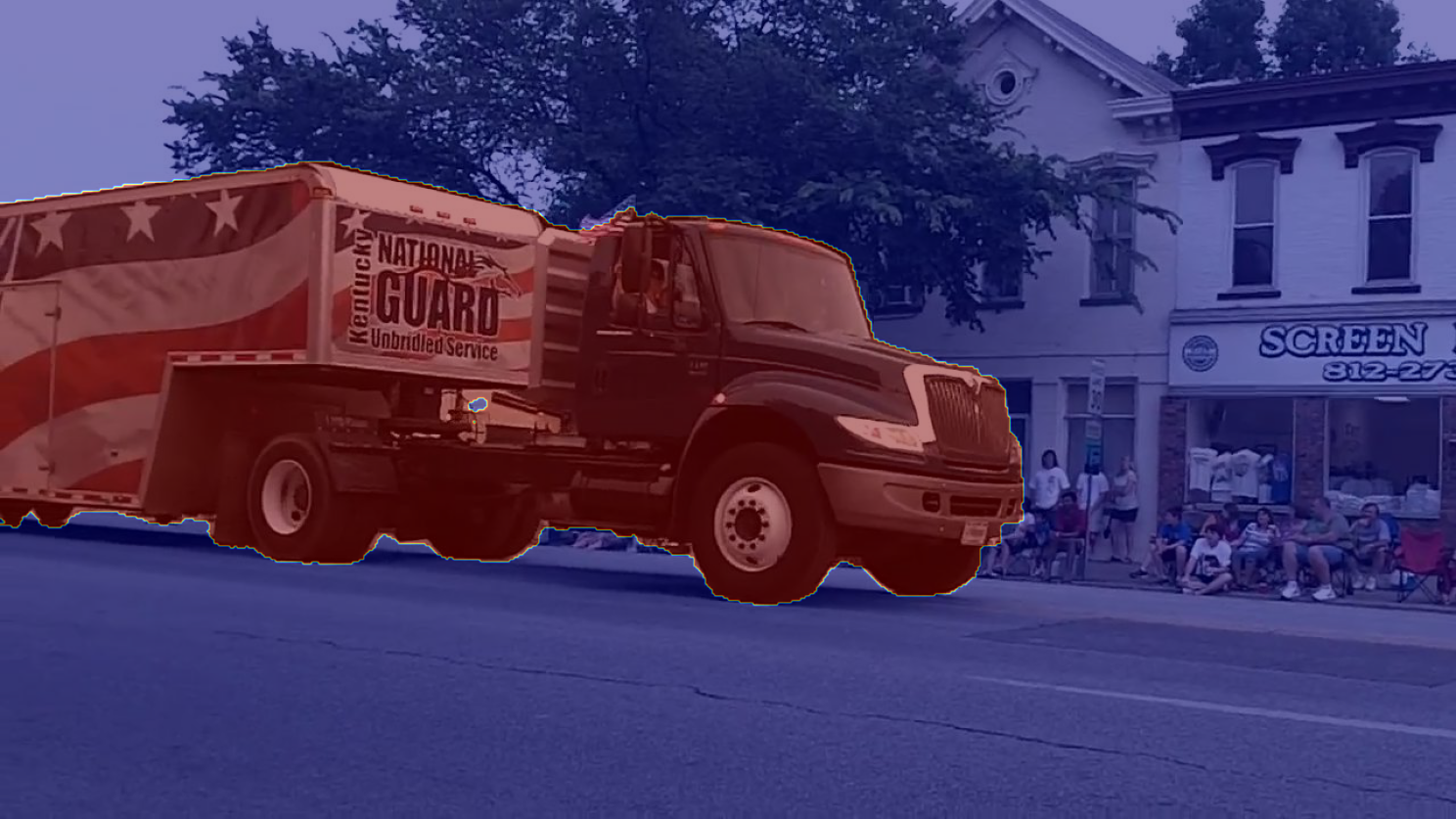}
        \includegraphics[width=0.22\linewidth]{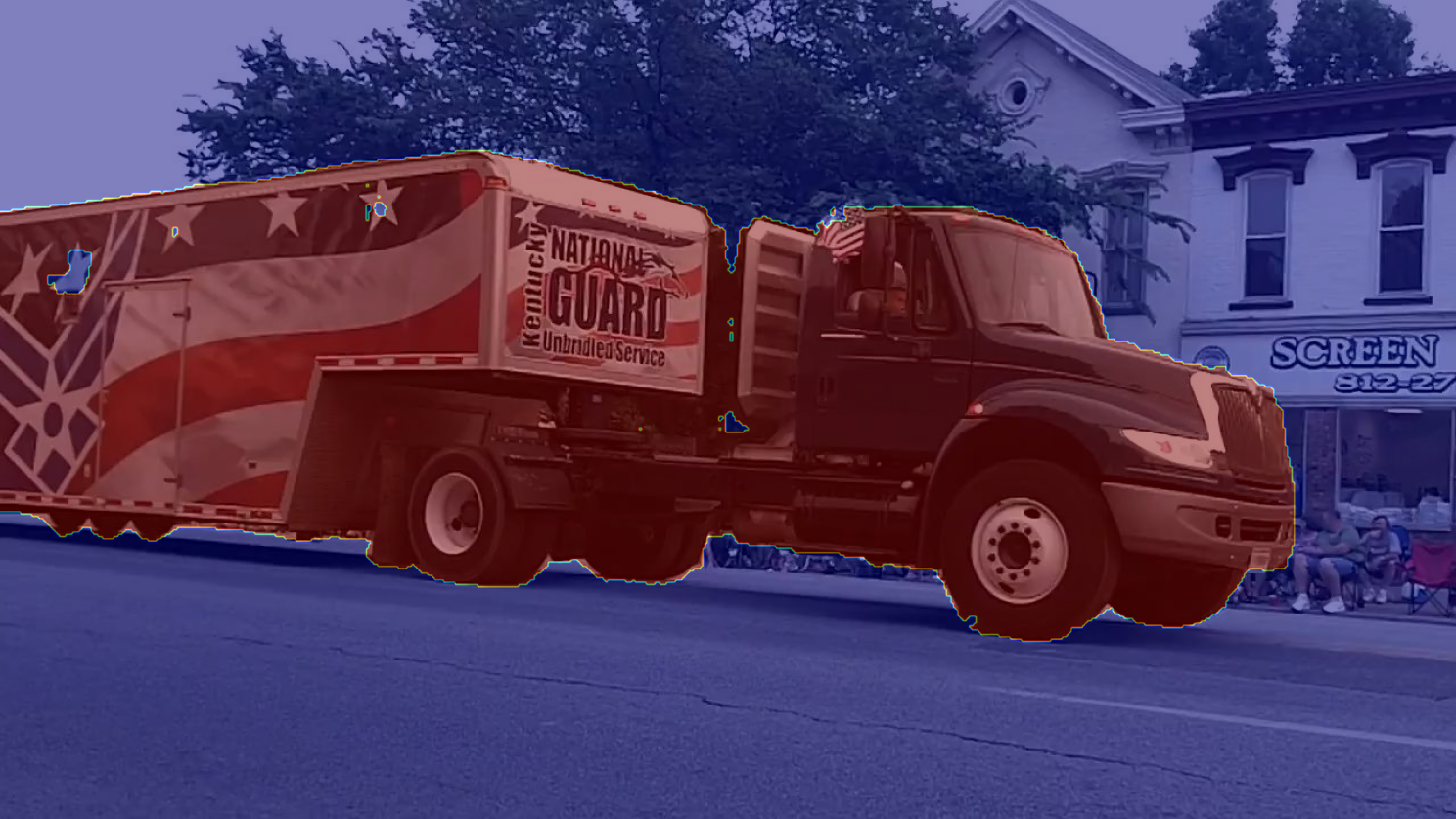}
        \caption{\textcolor{red}{the black truck with red/white and blue is moving down the road to the right with a crowd behind it}.}
        \label{fig:multiple_images53}
    \end{subfigure}
    
    \caption{Visualization of the impact of HDA in our RefSAM on Refer-Youtube-VOS. The first and second columns show the cases without using HDA, and the third and fourth columns are the showcases using HDA.  }
    \label{fig:multi_scale_visualization}
\end{figure*}

\begin{figure*}[!ht]
    \centering
    
    \hspace{1cm}
    \begin{subfigure}{\textwidth}
        \centering
        \includegraphics[width=0.22\linewidth]{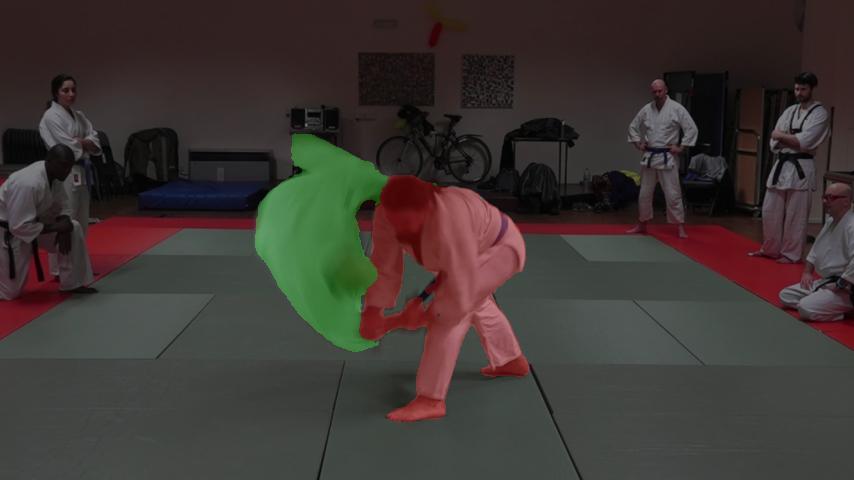}
        \includegraphics[width=0.22\linewidth]{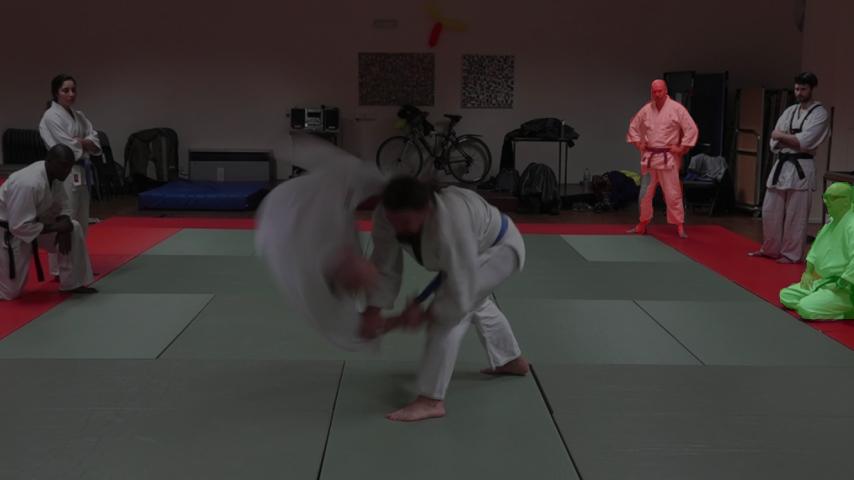}
        \includegraphics[width=0.22\linewidth]{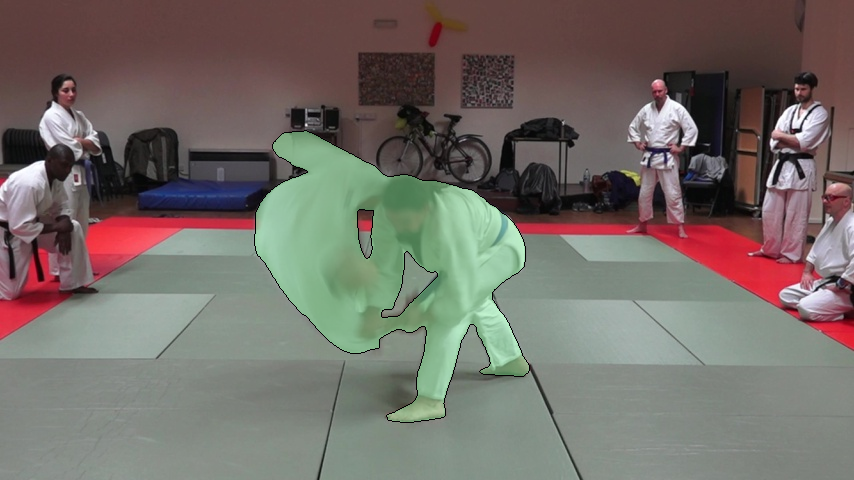}
        \includegraphics[width=0.22\linewidth]{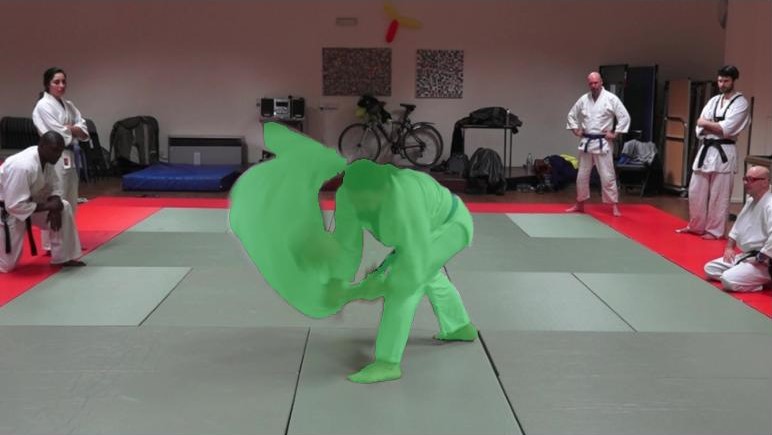}
        \caption{\textcolor{red}{A man with a blue belt on the right}. \textcolor{green}{A bald man with a black belt in the center}.}
        \label{fig:multiple_images41}
    \end{subfigure}

    \hspace{1cm}
    \begin{subfigure}{\textwidth}
        \centering
        \includegraphics[width=0.22\linewidth]{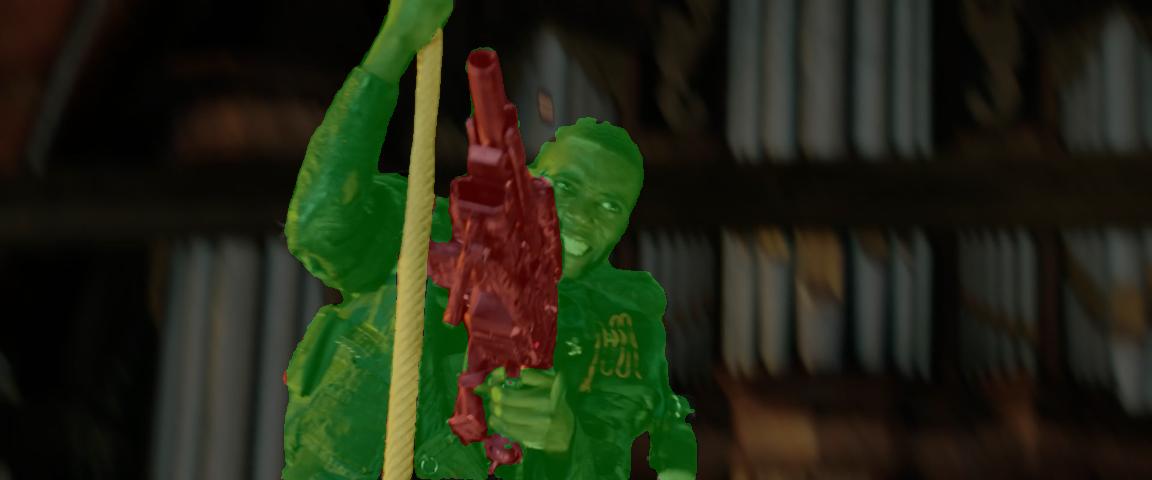}
        \includegraphics[width=0.22\linewidth]{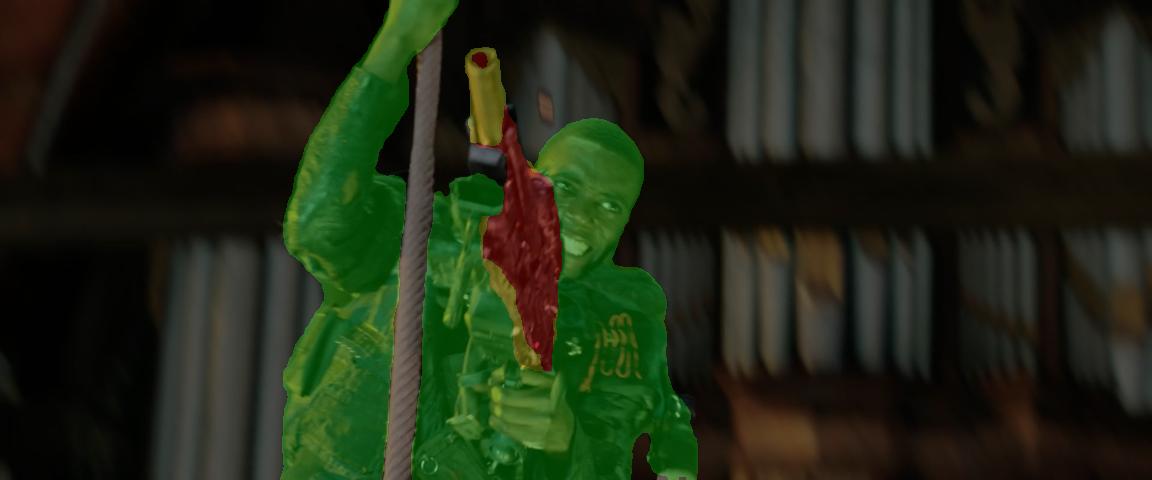}
        \includegraphics[width=0.22\linewidth]{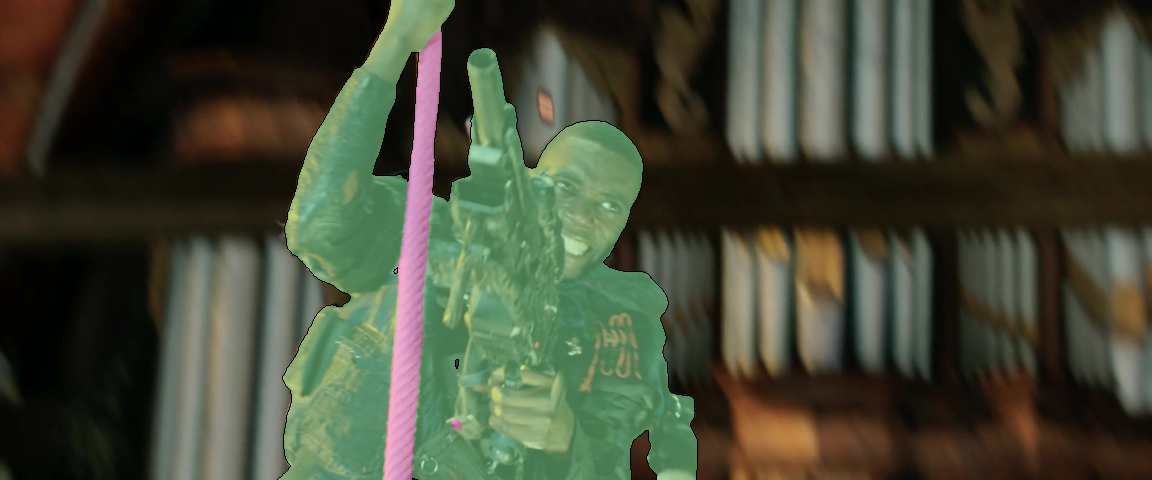}
        \includegraphics[width=0.22\linewidth]{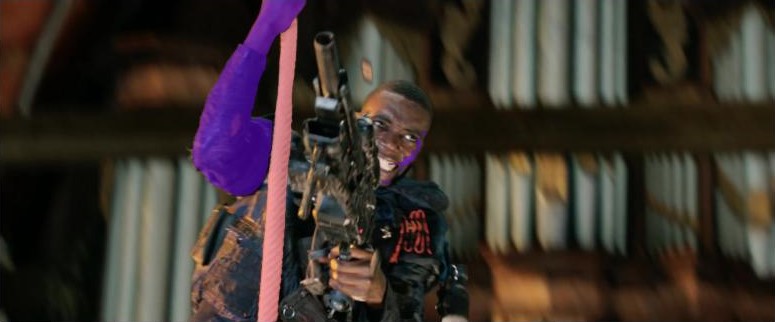}
        \caption{\textcolor{red}{A black shotting gun}. \textcolor{green}{A black man}. \textcolor{yellow}{A rope}.}
        \label{fig:multiple_images42}
    \end{subfigure}

    \hspace{1cm}
    \begin{subfigure}{\textwidth}
        \centering
        \includegraphics[width=0.22\linewidth]{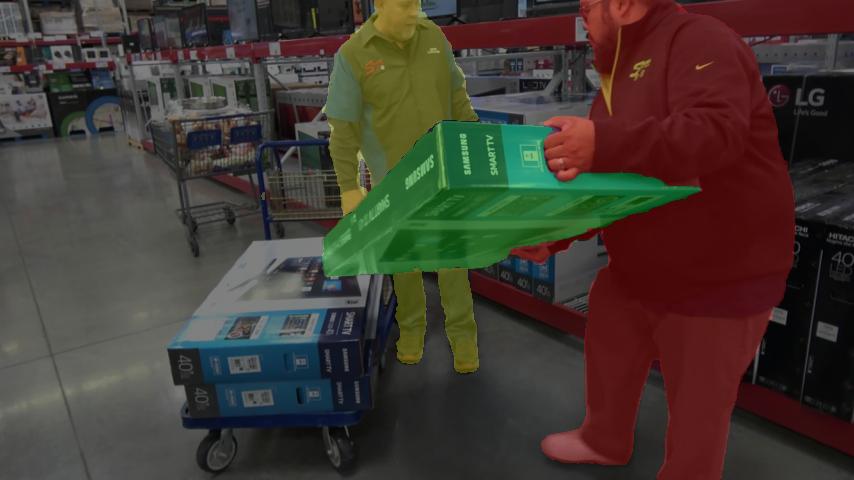}
        \includegraphics[width=0.22\linewidth]{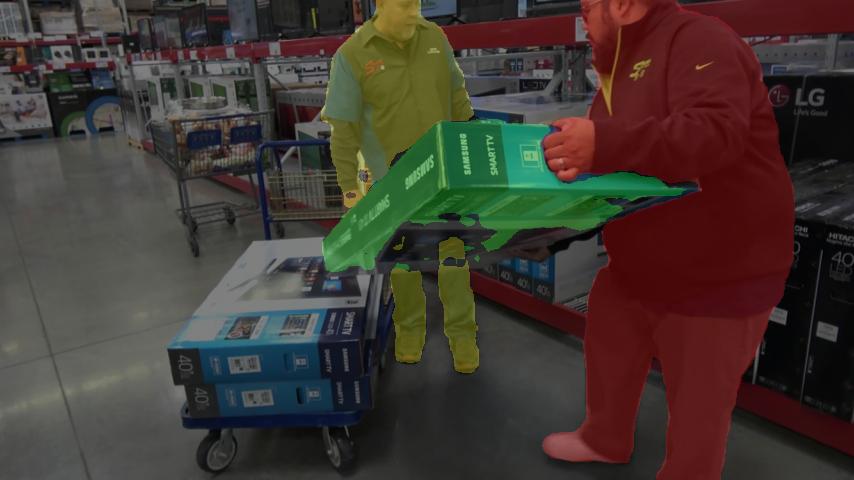}
        \includegraphics[width=0.22\linewidth]
        {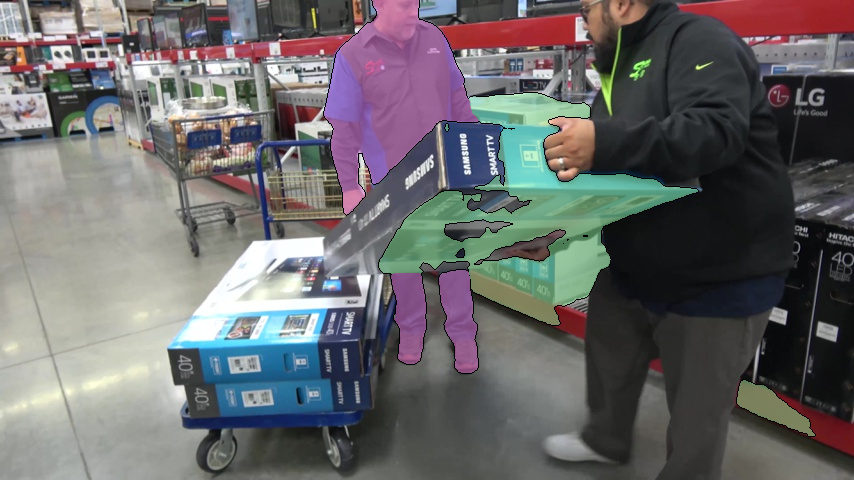}
        \includegraphics[width=0.22\linewidth]
        {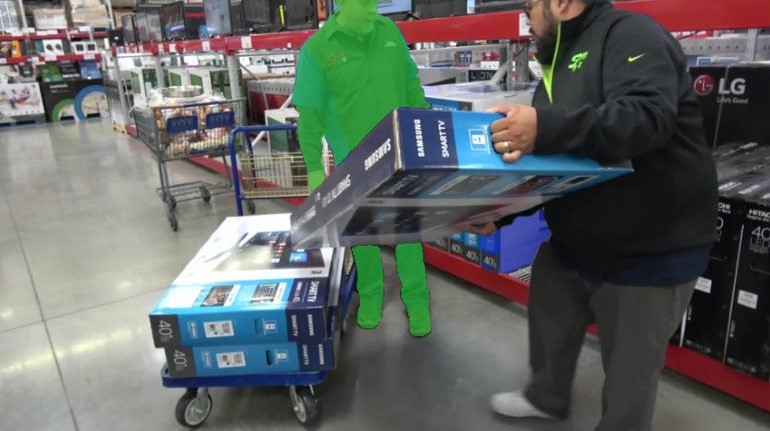}
        \caption{\textcolor{red}{A fat man on the right in a black jacket}. \textcolor{green}{A cardboard box held by a man}. \textcolor{yellow}{A man on the left with a beard wearing jeans}.}
        \label{fig:multiple_images43}
    \end{subfigure}
    
    \caption{Visualization of the results of different models on Ref-DAVIS17. From left to right: our RefSAM, ReferFormer \cite{wu2022language}, SAM-Track \cite{cheng2023segment} + Ground DINO \cite{liu2023grounding}, and PerSAM \cite{zhang2023personalize} + Ground DINO. The text with a specific color corresponds to the object mask of the same color.}
    \label{fig:mainfig4}
\end{figure*}

\subsubsection{Influence of Model Size}
\label{sec:Model Scale Analysis Experiment}
We then investigate the influence of model size, specifically the performance of using different sizes of Visual Encoder, including ViT-B, ViT-L, and ViT-H. All the models employ the Cross-modal MLP, HDA, Adapter, and ITM. The data augmentation described in Sec.~\ref{subsubsec:trainingdetails} is also used.The results on Refer-Youtube-VOS and Ref-DAVIS17 are shown in Table \ref{tbl3}. The findings indicate that employing a larger backbone for visual feature extraction yields superior performance compared to a smaller one, attributed to the enhanced representation capacity of larger vision transformers. Furthermore, the results exhibit a consistent improvement in performance with increasing model size, highlighting the good scalability of our RefSAM model.

\begin{table*}[!ht]
\caption{Ablation study of the Visual Encoder.} \label{tbl3}
    \centering 
    \begin{tabular}{cccccccc}
        \hline
            \multirow{2}{*}{Visual Encoder} & \multicolumn{3}{c}{Refer-Youtube-VOS} & \multicolumn{3}{c}{Ref-DAVIS17} \\
            \cmidrule(lr){2-4} \cmidrule(lr){5-7} 
            &  $\mathcal{J}$ \& $\mathcal{F}$ & $\mathcal{J}$ & $\mathcal{F}$ & $\mathcal{J}$ \& $\mathcal{F}$ & $\mathcal{J}$ & $\mathcal{F}$ \\ 
        \hline
            ViT-B & 58.4 & 57.4 & 59.4 & 62.1 & 59.0 & 65.3 \\
            ViT-L & 62.2 & 60.8 & 63.6 & 70.5 & 66.7 & 74.3 \\ 
            ViT-H & \textbf{67.6} & \textbf{65.8} & \textbf{69.4} & \textbf{71.9} & \textbf{68.4} & \textbf{75.5} \\      
        \hline
    \end{tabular}
\end{table*}

\subsection{Visualization Results}

We show the visualization results of our RefSAM model in Figure \ref{fig:visualresultsRefSAM}. It can be seen that RefSAM is capable of effectively segmenting and tracking the referred object even in challenging scenarios, such as variations in person poses, and occlusions between instances. Besides, we also visualize the impact of using HDA in Figure~\ref{fig:multi_scale_visualization}. It can be observed that leveraging HDA greatly enhances the model's capabilities in handling tiny objects and occlusions and providing more accurate segmentation masks, especially when dealing with complex text prompts. Furthermore, we present the results of different models in Figure \ref{fig:mainfig4}. It is clear that our RefSAM demonstrates significantly enhanced cross-modal understanding capability, particularly evident in handling vague language descriptions (Figure~\ref{fig:multiple_images41}) and resolving appearance ambiguity among similar objects (Figure~\ref{fig:multiple_images43}).

\begin{table*}[!ht]
\caption{Results on three representative RIS datasets.} \label{results_on_RIS_datasets}
    \centering
    \resizebox{\textwidth}{!}{ 
    \begin{tabular}{lllcccccccc}
    \hline
        \multirow{2}{*}{Methods} & \multirow{2}{*}{Visual Backbone} & \multirow{2}{*}{Textual Encoder} & \multicolumn{3}{c}{RefCOCO} & \multicolumn{3}{c}{RefCOCO+} & \multicolumn{2}{c}{G-Ref} \\
        \cmidrule(lr){4-6} \cmidrule(lr){7-9} \cmidrule(lr){10-11}
        &  &  & Val & Test A & Test B & Val & Test A & Test B & Val & Test \\ 
    \hline
        \noalign{\vskip 1pt} 
        \multicolumn{11}{l}{oIoU} \\ 
    \hline
    \noalign{\vskip 1pt} 
      CGAN~\cite{luo2020cascade}    & DeepLab-R101 & Bi-GRU &64.86 & 68.04 & 62.07 & 51.03 & 55.51 & 44.06 & 51.01 & 51.69 \\
      ISFP~\cite{liu2022instance}    & Darknet53 & Bi-GRU & 65.19 & 68.45 & 62.73 & 52.70 & 56.77 & 46.39 & 52.67 & 53.00 \\
      LTS~\cite{jing2021locate}     & Darknet53 & Bi-GRU & 65.43 & 67.76 & 63.08 & 54.21 & 58.32 & 48.02 & 54.40 & 54.25 \\     
      ReSTR~\cite{kim2022restr}   & ViT-B & TX & 67.22 & 69.30 & 64.45 & 55.78 & 60.44 & 48.27 & - & -  \\
      LAVT~\cite{yang2022lavt}    & Swin-B & BERT & 72.73 & 75.82 & 68.79 & 62.14 & 68.38 & 55.10 & 61.24 & 62.09 \\
      RefSegformer~\cite{wu2024towards} & Swin-B & BERT & 73.22 & 75.64 & 70.09 & 63.50 & 68.69 & 55.44 & 62.56 & 63.07 \\ 
    \hline
        \noalign{\vskip 1pt} 
        RefSAM & ViT-H & T5-3b & 74.89 & 77.49 & 70.90 & 65.88 & 71.60 & 57.26 & 66.46 & 67.23 \\ 
    \hline
        \noalign{\vskip 1pt} 
        \multicolumn{11}{l}{mIoU} \\ 
    \hline
      \noalign{\vskip 1pt} 
      VLT~\cite{ding2021vision}   & Darknet53 & Bi-GRU & 65.65 & 68.29 & 62.73 & 55.50 & 59.20 & 49.36 & 52.99 & 56.65 \\ 
      CRIS~\cite{wang2022cris}  & CLIP-L & CLIP-L & 70.47 & 73.18 & 66.10 & 62.27 & 68.06 & 53.68 & 59.87 & 60.36 \\ 
      SeqTR~\cite{zhu2022seqtr} & Darknet53 & Bi-GRU & 71.70 & 73.31 & 69.82 & 63.04 & 66.73 & 58.97 & 64.69 & 65.74 \\
      RefTR~\cite{li2021referring} & ResNet101 & BERT & 74.34 &76.77 & 70.87 & 66.75 & 70.58 & 59.40 & 66.63 & 67.39 \\
      LAVT~\cite{yang2022lavt}  & Swin-B & BERT & {74.46} & {76.89} & {70.94} & {65.81} & {70.97} & {59.23} & {63.34} & {63.62} \\
    \hline
        \noalign{\vskip 1pt} 
        RefSAM & ViT-H & T5-3b & 78.29 & 80.65 & 75.07 & 71.58 & 76.03 & 64.35 & 71.45 & 71.98 \\ 
    \hline
    \end{tabular}
    }
\end{table*}

\subsection{Model Complexity Analysis}
In this section, we perform an analysis of the complexity of various models. As depicted in Table~\ref{tbl3-Learnable-parameters}, our model exhibits a notably smaller number of learnable parameters in comparison to all other models, highlighting the inherent nature of parameter-efficient tuning employed during the training of our RefSAM model. Furthermore, Table~\ref{tbl3-Inference-time} presents the inference speed of various models. It is apparent that our RefSAM exhibits a marginally slower runtime compared to ReferFormer and SgMg, yet outperforms SAM-Track and PerSAM in terms of inference speed. When employing a ViT-B as the visual encoder, RefSAM achieves an inference speed of 9.5 FPS. 

\begin{table}[!ht]
\caption{The number of learnable parameters of different models.} \label{tbl3-Learnable-parameters}
    \centering
    \begin{tabular}{ccc}
    \hline
        Method & Visual Encoder & \#Params (M) \\ 
    \hline
        SAM & ViT-B & 93.7 \\
        SAM & ViT-L & 312.3 \\
        SAM & ViT-H & 641.1 \\
    \hline
        ReferFormer & Video-Swin-B & 112.9 \\
        ReferFormer & Swin-L & 221.7 \\
    \hline
        RefSAM & ViT-B &  9.9  \\ 
        RefSAM & ViT-L &  19.1 \\
        RefSAM & ViT-H &  33.0 \\
    \hline
    \end{tabular}
\end{table}

\begin{table}[!ht]
\caption{Inference speed of different models.} \label{tbl3-Inference-time}
    \centering
    \begin{tabular}{ccc}
    \hline
        Method & Backbone & FPS \\ 
    \hline
        OnlineRefer & Swin-L & 8.4 \\
        ReferFormer & Swin-L & 9.1 \\
        SgMg & Swin-L & 13.6 \\
    \hline
        SAM-Track + Ground DINO & ViT-H & 1.5 \\
        PerSAM + Ground DINO & ViT-H & 4.0 \\
    \hline
        RefSAM & ViT-H & 4.8 \\
        RefSAM & ViT-L & 6.5 \\
        RefSAM & ViT-B & 9.5 \\
    \hline
    \end{tabular}
\end{table}

\subsection{Results on RIS Datasets}
Our model can also be utilized for RIS. We evaluate our model on three representative datasets: RefCOCO \cite{yu2016modeling}, RefCOCO+ \cite{yu2016modeling} and G-Ref \cite{nagaraja2016modeling, mao2016generation}. RefCOCO comprises 142K referring language expressions describing 50K objects in approximately 20K images, while RefCOCO+ consists of 141K referring language expressions for 50K objects in roughly 20K images. G-Ref includes 104K referring language expressions for around 55K objects in about 27K images. 
Following ~\cite{yang2022lavt}, we train our model for 40 epochs using a batch size of 16 for each dataset. We use mean Intersection-over-Union (mIoU) and overall Intersection-over-Union (oIoU) as the evaluation metrics. For all experiments on the RIS datasets, we employ Cross-modal MLP, HDA and Adapter, and use the data augmentation described in Sec.~\ref{subsubsec:trainingdetails}. The results are shown in Table~\ref{results_on_RIS_datasets}. The proposed RefSAM achieves competitive results in terms of oIoU and mIoU. This highlights the versatility of our RefSAM and its exceptional performance for RIS.

\section{Conclusion}\label{sec5}
This study pioneers the adaptation of the foundational segmentation model, Segment Anything Model, to the referring video object segmentation task. We propose the novel RefSAM model, which incorporates efficient designs to effectively bridge the semantic gap between the visual and language domains, significantly enhancing the cross-modal understanding capability of SAM. By employing a parameter-efficient tuning strategy, RefSAM achieves efficient training through adjustments to a small number of learnable parameters. Extensive experiments on two representative referring video object segmentation datasets including Refer-Youtube-VOS and Ref-DAVIS17 datasets, as well as three referring image segmentation datasets validate the superior effectiveness of our RefSAM compared to existing methods.

\section{Data availability statements}\label{sec6}

Refer-Youtube-VOS is available for download from \emph{The Large-scale Video Object Segmentation Challenge} \footnote{\href{https://youtube-vos.org/}{https://youtube-vos.org/}} at \href{https://doi.org/10.1007/978-3-030-58555-6_13}{https://doi.org/10.1007/978-3-030-58555-6\_13}\cite{seo2020urvos}. 
Ref-DAVIS17 is available for download from \emph{International Max Planck Research School on Trustworthy Computing (IMPRS-TRUST)} \footnote{\href{https://www.mpi-inf.mpg.de/departments/computer-vision-and-machine-learning/research/video-segmentation/video-object-segmentation-with-language-referring-expressions}{https://www.mpi-inf.mpg.de/departments/computer-vision-and-machine-learning/research/video-segmentation/video-object-segmentation-with-language-referring-expressions}} at \href{https://doi.org/10.1007/978-3-030-20870-7_8}{https://doi.org/10.1007/978-3-030-20870-7\_8}\cite{khoreva2019video}.
RefCOCO and RefCOCO+ are available for download from \emph{UNC Vision Recognition Group} at \href{https://doi.org/10.1007/978-3-319-46475-6_5}{https://doi.org/10.1007/978-3-319-46475-6\_5}\cite{yu2016modeling}. 
G-Ref is available for download from \emph{UNC Vision Recognition Group} at \href{https://doi.org/10.1109/cvpr.2016.9}{https://doi.org/10.1109/cvpr.2016.9}\cite{mao2016generation}.


\bibliographystyle{IEEEtran}
\bibliography{IEEEabrv,IEEE}

\end{document}